\documentclass{article}
\usepackage{ijcai24}

\usepackage{times}
\usepackage{soul}
\usepackage{url}
\usepackage[hidelinks]{hyperref}
\usepackage[utf8]{inputenc}
\usepackage[small]{caption}
\usepackage{graphicx}
\usepackage{amsmath}
\usepackage{amsthm}
\usepackage{booktabs}
\usepackage{algorithm}
\usepackage{algorithmic}
\usepackage[switch]{lineno}

\usepackage{amsmath}
\usepackage{amssymb}
\usepackage{subcaption}
\usepackage{booktabs}
\usepackage{anyfontsize}

\title{ESS-ReduNet: Enhancing Subspace Separability of ReduNet via Dynamic Expansion with Bayesian Inference}

\author{
Xiaojie Yu$^1$
\and
Haibo Zhang$^1$\and
Lizhi Peng$^{2,3}$\and
Fengyang Sun$^4$\and
Jeremiah Deng$^1$\\
\affiliations
$^1$University of Otago\\
$^2$Quan Cheng Laboratory\\
$^3$University of Jinan\\
$^4$Victoria University of Wellington\\
\emails
yuxiaojie814@gmail.com,
haibo.zhang@otago.ac.nz,
plz@ujn.edu.cn,
fengyang.sun@vuw.ac.nz,
jeremiah.deng@otago.ac.nz
}

\begin{document}

\maketitle

\begin{abstract}
ReduNet is a deep neural network model that leverages the principle of maximal coding rate \textbf{redu}ction to transform original data samples into a low-dimensional, linear discriminative feature representation.
Unlike traditional deep learning frameworks, ReduNet constructs its parameters explicitly layer by layer, with each layer’s parameters derived based on the features transformed from the preceding layer. 
Rather than directly using labels,
ReduNet uses the similarity between each category’s spanned subspace and the data samples for feature updates at each layer.  
This may lead to features being updated in the wrong direction, impairing the correct construction of network parameters and reducing the network’s convergence speed. 
To address this issue, based on the geometric interpretation of the network parameters, this paper presents ESS-ReduNet to enhance the separability of each category’s subspace by dynamically controlling the expansion of the overall spanned space of the samples.
Meanwhile, label knowledge is incorporated with Bayesian inference to encourage the decoupling of subspaces.
Finally, stability, as assessed by the condition number, serves as an auxiliary criterion for halting training.
Experiments on the ESR, HAR, Covertype, and Gas datasets demonstrate that ESS-ReduNet achieves more than 10x improvement in  convergence compared to ReduNet. Notably, on the ESR dataset, the features transformed by ESS-ReduNet achieve a 47\% improvement in SVM classification accuracy.
\end{abstract}

\section{Introduction}\label{sec:intro}
For a classification task, deep neural networks are designed to learn a nonlinear mapping through a sequence of layers, with the goal of accurately mapping data to their respective labels. A common practice in training deep learning models involves minimizing empirical risk by employing cross-entropy (CE) loss \cite{goodfellow_deep_2016}. While CE loss is both effective and commonly used, fitting labels alone does not ensure learning meaningful, structured representational information. In fact, recent studies \cite{papyan_prevalence_2020,fang_exploring_2021,zhu_geometric_2021}
show that the learned representations derived from the training using CE loss demonstrate a phenomenon of \textit{neural collapse}: as the CE loss for every class gets close to zero, the representation of each class shrinks to a single point. The variability and structural details associated with each class are being stifled and ignored.
Additionally, the development of deep network architectures often stems from extensive trial and error, lacking a solid basis in explicit mathematical principles.

To address the dual questions concerning the objectives of representation learning and the principles of network architecture design, \citeauthor{chan_redunet_2022} \shortcite{chan_redunet_2022} introduced ReduNet, 
based on the lossy data coding and compression principles.
They proposed that the objective function for representation learning should focus on extracting \textbf{low-dimensional, linear discriminative representation} from high-dimensional data.
To learn linear discriminative representation, 
the following three principles are proposed:
1) intra-class compressible;
2) inter-class discriminative;
3) diverse representation where feature dimensions or variance within each class are maximized while remaining uncorrelated with features of other classes.
The quality of this type of representation can be measured using a principled metric derived from the lossy data compression, termed \textit{\textbf{rate reduction}}.
The architecture of ReduNet is constructed layer by layer in a forward fashion, with the goal of maximizing the \textit{rate reduction}. 
The parameters of each layer are constructed by the features transformed from the previous layer. 
Ideally, we expect the features of each class to update toward the subspaces spanned by the data of the corresponding class, with the subspaces becoming increasingly separated until they are mutually orthogonal.

\begin{figure}[t]
  \centering
    \includegraphics[width=0.80\linewidth]{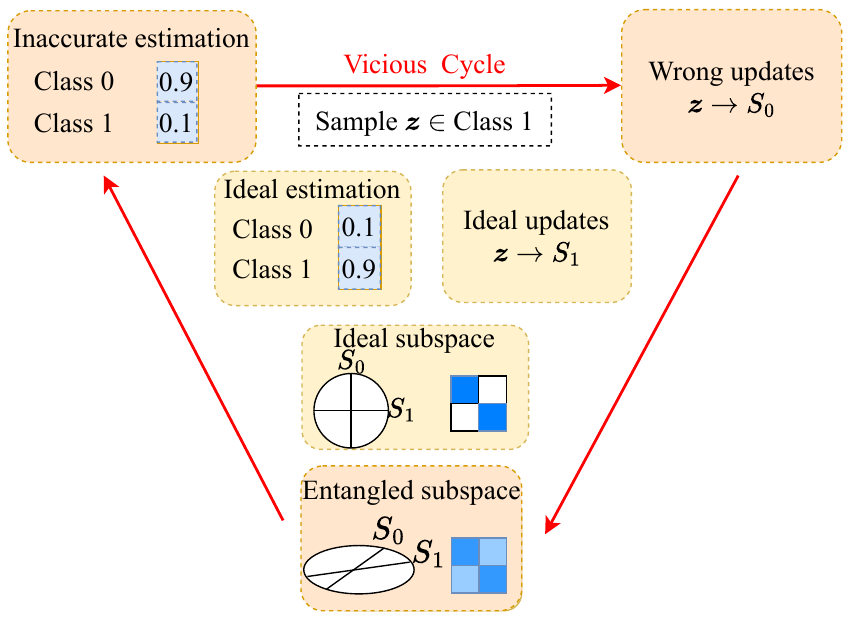}
  \caption{The Vicious Cycle of ReduNet. $S_0$ and $S_1$  are spanned subspaces of class $0$ and class $1$, respectively. $z \rightarrow S_0$ means sample $z$ updating towards $S_0$.}
  \label{fig:VCReduNet}
\end{figure}

Rather than directly using labels during training, ReduNet updates features based on estimated degree of membership by evaluating the similarities between the subspaces spanned by each class's data and the samples.
This helps to address the \textit{\textbf{inconsistency issue}}: Unlike the backpropagation strategy, the forward manner optimization strategy of ReduNet dictates that if each layer during training directly uses labels for feature updates, testing samples cannot be updated due to the lack of labels.
However, the estimated degree of membership can be highly inaccurate in the early layers of the network.
The inaccuracy of the estimation arises because the separability of subspaces may be compromised due to the limited overall spanned space.
The constrained and entangled subspaces may fail to accurately estimate 
the degree of membership effectively.
As illustrated in Figure \ref{fig:VCReduNet}, this leads to an incorrect update of features and inaccurate network parameters, further deteriorating the separability of the subspace and forming a vicious cycle.
This also decreases the network’s convergence speed and the final classification accuracy of the transformed features.
Besides,
ReduNet needs to carefully manage early stopping for its training efficiency and quality.
Firstly, if ReduNet is correctly constructed, the objective function stabilizes quickly, making additional layers unnecessary.
Secondly, although the objective function of an incorrectly constructed ReduNet tends to be stable, the vicious cycle will continue to deteriorate the quality of features and the network.
Therefore, relying solely on the objective function to stop training is unreasonable.
In this paper, we present ESS-ReduNet, a framework that aims to \textbf{e}nhance \textbf{s}ubspace \textbf{s}eparability to correct the feature updates and accelerate training with the following contributions:
\begin{enumerate}
  \item 
  To prioritize the expansion of the overall spanned space, we introduce a weight function to dynamically control the expansion process.
  This enhances the separability of subspaces across different classes in higher-dimensional space, thereby guiding the samples to update in the correct direction.
  \item 
  By comparing erroneous estimations with labels, the label knowledge is incorporated by Bayesian inference to correct the estimation of membership.
  Note that this introduces label knowledge while avoiding the inconsistency issue, as the posterior probabilities obtained through Bayesian inference during training retain the likelihood of errors in estimation functions, which can be reused during testing.
  \item 
  Given that ReduNet aims to flatten each category of data into its respective linear subspace, we use the condition number, an essential metric for testing whether a linear system is ill-conditioned, as an auxiliary criterion for stopping training.
  This helps save computational resources and maintain feature quality.
  \item Experiments on the HAR and ESR datasets demonstrate that our method accelerates the reduction of errors in estimating membership by a factor of 100 compared to ReduNet. Additionally, it speeds up network convergence by tenfold on the ESR, HAR, Covertype, and Gas datasets. Furthermore, on the ESR and Covertype datasets, the transformed features by our method achieved a 47\% and 37\% increase in SVM accuracy, respectively. Significant improvements were also observed in other datasets and with other classifiers (KNN, NSC).

\end{enumerate}


\section{Background}\label{sec:bg}
\subsection{Overview of ReduNet}
For the given finite data samples \( \boldsymbol{X} = [\boldsymbol{x}^1, \dots, \boldsymbol{x}^m] \in \mathbb{R}^{D \times m} \), the compactness of the learned features \( \boldsymbol{Z} = [\boldsymbol{z}^1, \dots, \boldsymbol{z}^m] \in \mathbb{R}^{n \times m}\) can be measured by the average coding length, i.e. the \textit{coding rate} subject to the distortion  \(\epsilon\)
:
\begin{equation}
  \label{eq:codingrate}
  R(\boldsymbol{Z},\epsilon)
  \overset{\cdot}{=} \frac{1}{2}\log \det(\boldsymbol{I}+\frac{n}{m\epsilon^2} \boldsymbol{Z}\boldsymbol{Z}^{T}).
\end{equation}
This measurement was proposed by \citeauthor{ma_segmentation_2007} \shortcite{ma_segmentation_2007} based on the rate-distortion theory
\cite{cover_elements_2006}.
Supposing $\boldsymbol{Z}$ contains 
two \textbf{uncorrelated} subsets, 
$\boldsymbol{Z}_1$ and $\boldsymbol{Z}_2$,
the coding rate for all data $R(\boldsymbol{Z}_1 \cup \boldsymbol{Z}_2)$ is greater than the sum of the coding rates for $R(\boldsymbol{Z}_1)$ and $R(\boldsymbol{Z}_2)$.
Furthermore,  if \( \boldsymbol{Z}_1\) is \textbf{orthogonal} to \( \boldsymbol{Z}_2\), the difference \(R(\boldsymbol{Z}_1 \cup \boldsymbol{Z}_2) - (R(\boldsymbol{Z}_1) + R(\boldsymbol{Z}_2)) \) is approximately maximized.

For a classification problem involving 
$k$ classes, \citeauthor{chan_redunet_2022} \shortcite{chan_redunet_2022} proposed ReduNet, which aims to perform classification by implementing a series of transformations that orthogonalize the samples across different classes.
Hence, 
\textit{maximal coding rate reduction} (\(\mathrm{MCR}^2\)) is proposed as the objective function:
\begin{equation}
  \label{eq:originobjectivefunction}
  \begin{split}
    \Delta R(\boldsymbol{Z},\mathbf{\Pi},\epsilon) &=
    R(\boldsymbol{Z},\epsilon) - R_c(\boldsymbol{Z}, \epsilon \mid \mathbf{\Pi}) \\
  \end{split}
\end{equation}
$R(\boldsymbol{Z},\epsilon)=\frac{1}{2}\log \det(\boldsymbol{I}+\alpha \boldsymbol{Z}\boldsymbol{Z}^{T})$ aims to maximize the spanned space volume (or dimension) 
of the whole set \( \boldsymbol{Z}\) such that features of different classes are incoherent to each other (i.e. inter-class discriminative). 
$- R_c(\boldsymbol{Z}, \epsilon \mid \mathbf{\Pi})= - \sum_{j=1}^{k}\frac{\gamma_j}{2}\log \det(\boldsymbol{I}+\alpha_j\boldsymbol{Z}\mathbf{\Pi}^{j}\boldsymbol{Z}^{T})$ aims to minimize the spanned space volume
of each class such that features within the same class are highly correlated (i.e. intra-class compressible).
Here, \(\mathbf{\Pi} = \{ \mathbf{\Pi}^j \in \mathbb{R}^{m \times m}\}_{j=1}^k\) is defined as a group of diagonal matrices, which are used to encode the membership of  \(m\) samples within \(k\) different classes. Specifically, \(\mathbf{\Pi}^j(i,i)\) represents the probability of the \(i^{th}\) sample belonging to  \(j^{th}\) subset. Similar to the denotation in \cite{chan_redunet_2022},  \( \alpha = \frac{n}{m\epsilon^2}\), \(\alpha_j = \frac{n}{\mathrm{tr}(\mathbf{\Pi}^j) \epsilon^2}\),\(\gamma_j = \frac{\mathrm{tr}(\mathbf{\Pi}^j)}{m}\) for \(j = 1,\dots,k\).

\begin{figure}[t]
  \centering
    \includegraphics[width=0.95\linewidth]{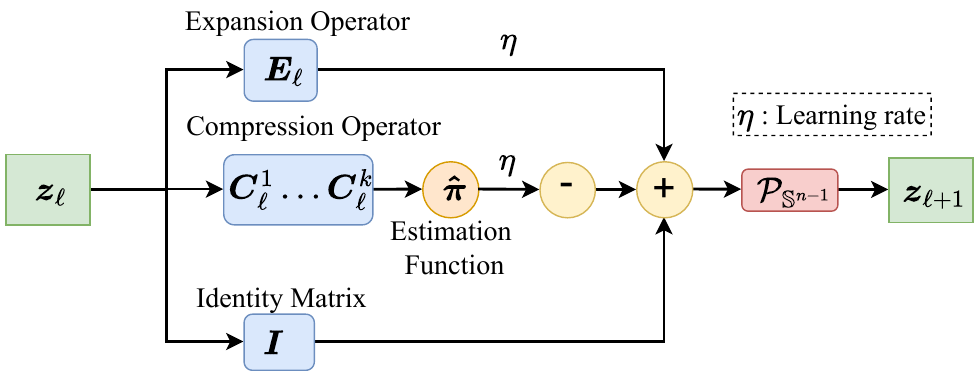}
  \caption{A Layer of ReduNet}
  \label{fig:ReduNet}
\end{figure}
To maximize \(\mathrm{MCR}^2\), a gradient ascent method based on the derivatives of the objective function Eq. \eqref{eq:originobjectivefunction} is used for feature updates:
\begin{equation}
  \label{eq:partialOfOrigin}
  \left. \frac{\partial\Delta R}{\partial \boldsymbol{Z}} \right|_{\boldsymbol{Z}_{\ell}} = \underbrace{\boldsymbol{E}_{\ell}}_{\text{Expansion}}\boldsymbol{Z}_{\ell} - \sum_{j=1}^{k}\gamma_j \underbrace{\boldsymbol{C}^{j}_{\ell}}_{\text{Compression}}\boldsymbol{Z}_{\ell}\mathbf{\Pi}^{j}.
\end{equation}
Here, \(\boldsymbol{E}_{\ell} = \alpha (\boldsymbol{I}+\alpha\boldsymbol{Z}_{\ell}\boldsymbol{Z}^{T}_{\ell})^{-1} \in \mathbb{R}^{n \times n}\) and $\boldsymbol{C}^{j}_{\ell} = \alpha_j(\boldsymbol{I}+\alpha_j\boldsymbol{Z}_{\ell}\mathbf{\Pi}^{j}\boldsymbol{Z}^{T}_{\ell})^{-1} \in \mathbb{R}^{n \times n}$.
Thereby,
for a feature \(\boldsymbol{z}_{\ell}\), the increment transform $g(\cdot, \boldsymbol{\theta}_{\ell} )$ on the $\ell$-th layer is defined as:
\begin{equation}
    \boldsymbol{z}_{\ell+1} \varpropto \boldsymbol{z}_{\ell} + \eta \cdot g(\boldsymbol{z}_{\ell}, \boldsymbol{\theta}_{\ell}) 
    \quad  \text{subject to} \quad  \boldsymbol{z}_{\ell+1} \in \mathbb{S}^{n-1}.
\end{equation}
\begin{equation}
  \label{eq:incrementTransform}
  g(\boldsymbol{z}_{\ell}, \boldsymbol{\theta}_{\ell}) = \boldsymbol{E}_{\ell}\boldsymbol{z}_{\ell} - \sum_{j=1}^{k}\gamma_j \boldsymbol{C}^{j}_{\ell}\boldsymbol{z}_{\ell}\boldsymbol{\hat{\pi}}^{j} (\boldsymbol{z}_{\ell}) \in \mathbb{R}^{n}.
\end{equation}
Note that unlike Eq.\ref{eq:partialOfOrigin}, Eq.\ref{eq:incrementTransform} uses estimation functions $\{\boldsymbol{\hat{\pi}}^{j}\}_{j=1}^k$ for feature updates instead of directly using the labels. This allows us to consistently use the same estimation functions for feature updates during the testing, without worrying about the inconsistency issues that may arise from directly using the labels.
Hence, 
as shown in Figure \ref{fig:ReduNet},
ReduNet is built in a layer-by-layer forward manner. 
For the \(\ell\)-th layer, the parameters  (i.e., expansion operators \(\boldsymbol{E}_{\ell}\) and compression operators $\{\boldsymbol{C}_{\ell}^j\}^k_{j=1}$) are constructed by transformed features $\boldsymbol{Z}_{\ell}$ of previous layer.
All features are then updated by \( \boldsymbol{E}_{\ell} \) and $\{\boldsymbol{C}_{\ell}^j\}^k_{j=1}$.
Ultimately, the objective is to ensure that the transformed features from different classes are orthogonal to each other.

In practice, under the assumption that signals are sparsely generated, ReduNet introduces a \textbf{\textit{lifting}} operation, which involves convolving the samples with $N_c$ filters to obtain features across $N_c$ channels. 
This  \textit{lifting} operation
expands the upper limit of the spanned space's dimensions, transforming from the original $\mathbb{R}^{n \times m}$ to $\mathbb{R}^{N_c \times n \times m}$.
Nevertheless, for some challenging datasets, ReduNet fails to fully utilize the expanded space because the rank of the spanned space does not increase sufficiently (at most \( N_c \times n \times m \)), and as a result, it continues to transform features within a constrained, smaller space.
This hinders the sufficient  expansion of the subspaces for each class (i.e., the separation between subspaces), thereby affecting the accurate estimation of the degree of membership and leading to incorrect feature updates.

\subsection{Related Works}
\subsubsection{Efforts on Subspace Representation Learning}
A common belief is that the data from each class exhibit a low-dimensional intrinsic structure, and the role of deep networks is to learn these structures \cite{hinton_reducing_2006}. 
Although many efforts seek to directly impose subspace structures on features learned by deep networks \cite{ji_deep_2017,peng_deep_2017,lezama_ole_2018,zhou_deep_2018,jawahar_scalable_2019,zhang_neural_2019},
the subspace properties explored in these studies do not fulfill the three principles mentioned in section \ref{sec:intro} \cite{haeffele_critique_2021}. In contrast, the representation derived from the optimized $\mathrm{MCR}^2$ objective function exhibits the desired characteristics.
\subsubsection{Attempts on Architecture Explanation}
\citeauthor{chan_redunet_2022} \shortcite{chan_redunet_2022} also aim to explain components of deep networks within a unified theoretical framework, such as convolutions \cite{lecun_gradient-based_1998,krizhevsky_imagenet_2012}, skip connection \cite{he_deep_2016}, nonlinear activation functions, etc.
They established a connection between deep convolutional networks and invariant rate reduction. 
Using the relationship between the circulant matrix and the convolution operation, the deep network derived from $\mathrm{MCR}^2$ could naturally transition to a deep convolutional network. 
\textit{Sparse rate reduction} can also be used to construct white-box transformers, which bridge compresssion, denoising, and multi-head self-attention \cite{yu_white-box_2023,yu_white-box_2023-1}.
In addition, the relationship between the circulant matrix and the discrete Fourier transform can be utilized to develop a \textbf{Fourier version} of ReduNet.
Due to the slower convergence speed of the Fourier version of ReduNet, this paper primarily focuses on the basic version of ReduNet.
However, our \textbf{plug-in} approach is equally applicable to the Fourier version, as our improvements solely involve adjusting the weights of the expansion operators and correcting errors in estimating labels.


ReduNet suffers from the limitations of a low-volume spanned space and errors in estimating membership. 
We will discuss these issues in detail in the next section. 
Furthermore, we will show that 
relying solely on incorporating label knowledge to correct network training quickly leads to stagnation. 
Further expansion of the spanned space is necessary to correct and accelerate network training.
This work highlights the relative importance of expansion operations: \textit{ 
only a sufficiently large overall data span can support the compression and transformation of individual subspaces; otherwise, in particularly small spaces, subspaces can easily become entangled.}

\section{Motivation}
\label{sec:motivation}
To demonstrate the motivation, the Kaggle
version \shortcite{ur-rashid_epileptic_2018} of the ESR \cite{qiuyi_wu_epileptic_2017} dataset serves as a case study of the binary classification problem, distinguishing between instances with and without epileptic seizures. 
After balancing by random undersampling, each class containing 1,456 samples, with each sample having 178 dimensions.
\begin{figure}[t]
  \centering
  \begin{subfigure}{0.49\linewidth}
  \centering
    \includegraphics[width=\linewidth]{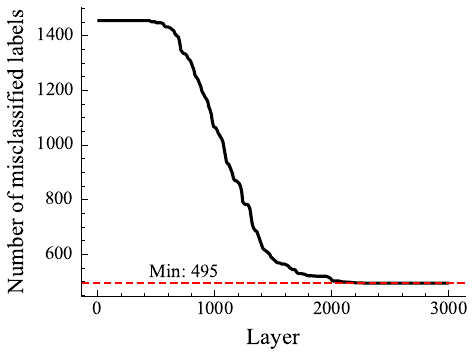}
    \caption{}
    \label{fig:originofYerr}
  \end{subfigure}
  \begin{subfigure}{0.49\linewidth}
  \centering
    \includegraphics[width=\linewidth]{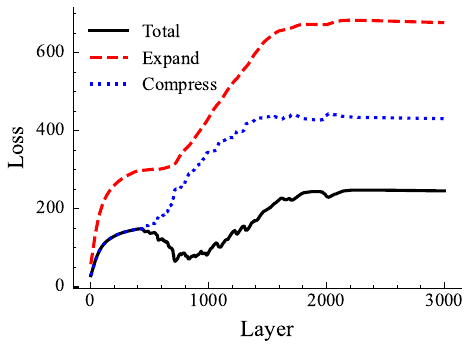}
    \caption{}
    \label{fig:MCRofOriginESR}
  \end{subfigure}
  
  \begin{subfigure}{0.49\linewidth}
  \centering\includegraphics[width=\linewidth]{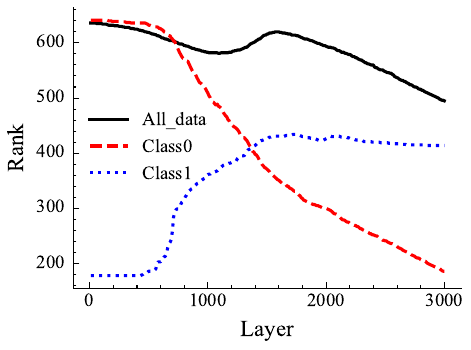}
    \caption{}
    \label{fig:RankofESR}
  \end{subfigure}
  \begin{subfigure}{0.49\linewidth}
  \centering
    \includegraphics[width=\linewidth]{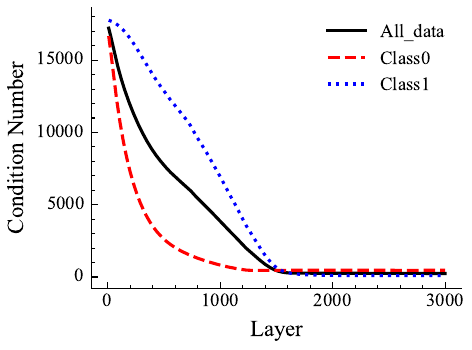}
    \caption{}
    \label{fig:ConditionNumberofESR}
  \end{subfigure}
  \caption{
  (a): The number of misclassified labels of \(\{\boldsymbol{\hat{{\pi}}^j}\}^k_{j=1}\)
; 
  (b): Objective function curve
; 
  (c): Rank trend
; 
  (d): Condition number trend. 
  }
  \label{fig:test}
\end{figure}

\textbf{Motivation 1:} \textit{constrained and entangled subspaces reduce the accuracy of estimation functions, thereby hindering the correct updating of features and the proper construction of the network parameters.}
As shown in Eq.\ref{eq:incrementTransform},
to ensure consistency between the training and testing,
ReduNet uses $k$ estimation functions \(\{\boldsymbol{\hat{{\pi}}^j}\}^k_{j=1}\) to approximate membership functions \(\{\boldsymbol{{\pi}}^j\}^k_{j=1}\),
where 
\begin{equation}
\boldsymbol{\hat{\pi}}^j(\boldsymbol{z}^i_{\ell}) \overset{\cdot}{=} \frac{\exp(-\lambda\|\boldsymbol{C}_{\ell}^{j}\boldsymbol{z}^i_{\ell}\|)}{\sum_{j=1}^{k}\exp(-\lambda\| \boldsymbol{C}_{\ell}^{j}\boldsymbol{z}^i_{\ell}\|)} \in [0,1];
\end{equation}
Essentially, these functions estimate membership of samples based on the similarity between the subspaces of the 
$k$ classes and the samples.
As the Figure \ref{fig:originofYerr} shows, although the misclassification rate of estimation functions decreases with more layers, these functions still can not  provide correct estimations for some challenging samples.
Besides,
Figure \ref{fig:MCRofOriginESR} shows the  curve of objective function \(\mathrm{MCR}^2\).
The \textit{Expand} curve corresponds to $R(\boldsymbol{Z},\epsilon)$ of Eq. \eqref{eq:originobjectivefunction}, while the \textit{Compress} curve corresponds to $ R_c(\boldsymbol{Z}, \epsilon \mid \mathbf{\Pi})$ of Eq. \eqref{eq:originobjectivefunction}. The \textit{Total} curve represents the value of the \(\mathrm{MCR}^2\) function. 
It can be observed that the \textit{Total} curve overlaps with the \textit{Compress} curve approximately before the 500th layer. 
We identified this as an extreme case of the estimation function's failure, referred to as the \textit{\textbf{lopsided}} issue. 
This occurs when samples from one class (Class 1) are incorrectly classified by the estimation function into another class (Class 0), resulting in erroneous updates for all samples of the misclassified class. 
Specifically,
samples from Class 1  are only compressed by the compression operator $\boldsymbol{C}^0_{\ell}$.
Furthermore, the network parameters constructed by the degraded feature are inappropriate.


\textbf{Motivation 2:} \textit{suboptimal estimation functions decrease the network's convergence speed, and the objective function should not be the sole criterion for assessing network convergence.} Figures \ref{fig:originofYerr} and \ref{fig:MCRofOriginESR} demonstrate that ReduNet requires more than 2000 layers to achieve convergence.
Nevertheless, the ranks of the data matrices continues to decline after 2000 layers, as illustrated in Figure \ref{fig:RankofESR}. 
This indicates that further training of a poorly constructed network degrades feature quality.
Inspired by \cite{curth_u-turn_2023},  the condition numbers of the matrices $(\boldsymbol{I}+\alpha\boldsymbol{Z}_{\ell}\boldsymbol{Z}^{*}_{\ell})$ and $\{(\boldsymbol{I}+\alpha_j\boldsymbol{Z}_{\ell}\mathbf{\Pi}^{j}\boldsymbol{Z}^{*}_{\ell})\}^k_{j=1}$ in the \( \boldsymbol{E}_{\ell} \) and $\{\boldsymbol{C}_{\ell}^j\}^k_{j=1}$ are measured, as depicted in Figure \ref{fig:ConditionNumberofESR}. 
The condition number stabilizes around the 1600th layer, earlier than the objective function at 2000 layers.
Hence, the changes of condition number can be an  auxiliary criterion to halt training. 
This helps save computational resources and maintain feature quality.
\section{ESS-ReduNet}
\label{sec:methods}
Three questions need to be addressed:
\begin{enumerate}
    \item Can we introduce labels to guide feature updates in a relatively correct direction without causing the inconsistency issue?
    \item  Besides using labels, are there other methods to improve the accuracy of estimation functions?
    \item Are there other metrics that can be used to assist in determining when to stop training?
\end{enumerate}
\begin{figure}[t]
  \centering
    \includegraphics[width=0.95\linewidth]{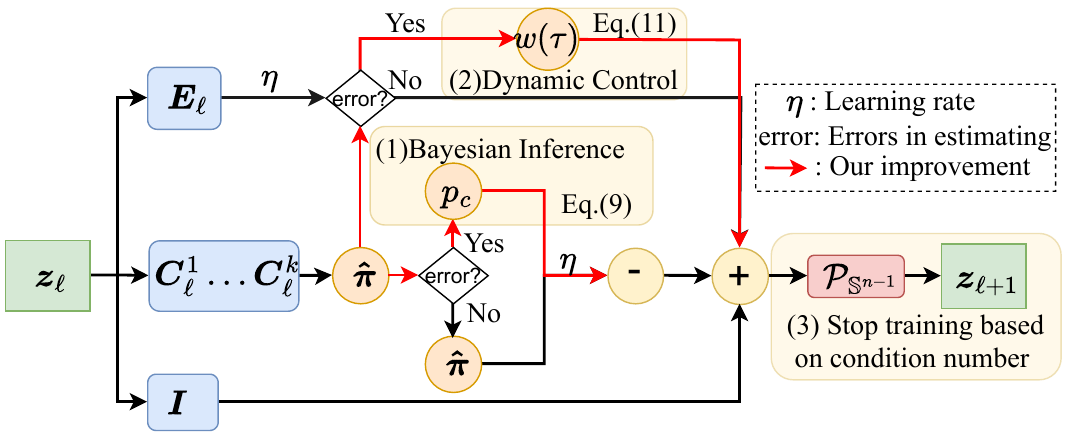}
  \caption{Overview of ESS-ReduNet}
  \label{fig:reviseReduNet}
\end{figure}

\subsection{Incorporate Label Knowledge
}
A straightforward approach to addressing the inaccuracy of the estimation function is to use labels for correcting the estimates.
However, this can easily introduce inconsistency issues once again.
What we aim for is to find a method that can both use label knowledge to guide feature updates and allow for the reuse of this information during the test phase.
By combining the labels with the estimates of the estimation function, we can calculate the average performance of the estimation function on samples from each class, and infer the posterior probability based on the labels: when a sample is observed to be classified into class \(i\), what is the probability that it actually belongs to class \(j\)? We can treat this posterior probability as containing certain label knowledge, which can be directly reused during the test phase.
Hence, Bayesian inference is employed to incorporate the label knowledge. 

The underlying intuition is to first count the samples misclassified by estimation functions, and then by comparing with the labels, infer the posterior probability that a sample observed to be classified into class $C_j$ (i.e. $\boldsymbol{z}_{\ell} \rightarrow C_j$) actually belongs to class $C_i$ (i.e. $\boldsymbol{z}_{\ell} \in C_i$): $p^{ij} = P(\boldsymbol{z}_{\ell} \in C_i | \boldsymbol{z}_{\ell} \rightarrow C_j )$.
\begin{equation}
  \label{eq:bayes}
  \begin{split}
    p^{ij}
    &= \frac{P(\boldsymbol{z}_{\ell} \in C_i)P(\boldsymbol{z}_{\ell} \rightarrow C_j | \boldsymbol{z}_{\ell} \in C_i)}{\sum_i P(\boldsymbol{z}_{\ell} \rightarrow C_j | \boldsymbol{z}_{\ell} \in C_i)P(\boldsymbol{z}_{\ell} \in C_i)}
  \end{split}
\end{equation}
Given that this study uses balanced datasets with $k$ classes, the prior probability $P(\boldsymbol{z}_{\ell} \in C_i) = \frac{1}{k}$.
For samples of class $i$, the probability of being classified into class $j$, i.e. $P(\boldsymbol{z}_{\ell} \rightarrow C_j | \boldsymbol{z}_{\ell} \in C_i)$ in Eq.\ref{eq:bayes}, is defined as the average probability of all samples in class $i$ being assigned by the estimation function $\boldsymbol{\hat{{\pi}}}^j$ during the training process:
\begin{equation}
  \label{eq:aveProb}
    P(\boldsymbol{z}_{\ell} \rightarrow C_j | \boldsymbol{z}_{\ell} \in C_i) = \frac{1}{m_i}\sum_{l=1}^{m_i} \boldsymbol{\hat{\pi}}^j(\boldsymbol{z}_{\ell}^l)
\end{equation}
Here, $m_i$ is the number of samples in class $i$. 
Therefore, for a sample $\boldsymbol{z}_{\ell}$, we use the corrected estimation $p_c^i(\boldsymbol{z}_{\ell}) = P_c(\boldsymbol{z}_{\ell} \in C_i)$ as the weight for the compression operator $\boldsymbol{C}_{\ell}^i$:
\begin{equation}
  \label{eq:Probofz}
  \begin{split}
    p_c^i(\boldsymbol{z}_{\ell})
    &= 
    \sum_{j=1}^{k}  P(\boldsymbol{z}_{\ell} \in C_i | \boldsymbol{z}_{\ell} \rightarrow C_j ) P(\boldsymbol{z}_{\ell} \rightarrow C_j) \\
    \end{split}
\end{equation}
Here, $P(\boldsymbol{z}_{\ell} \rightarrow C_j) =\boldsymbol{\hat{\pi}}^j(\boldsymbol{z}_{\ell})$ represents the estimated probability that the sample $\boldsymbol{z}_{\ell}$ belongs to class $j$, as derived from the estimation function.

Although Bayesian inference can correct the estimation ideally, the overall spanned space may not expand promptly, preventing further decoupling of the subspaces and leading to stagnation of the training process. Figure \ref{fig:YerrBayesOrigin} from the ablation study confirms this assumption. Consequently, dynamic control of the expansion is introduced next.

\subsection{Control the Expansion Dynamically}
\subsubsection{Analysis of the Geometric Interpretation of Expansion Operators}
\begin{figure}[hbt]
  \centering
    \includegraphics[width=0.45\linewidth]{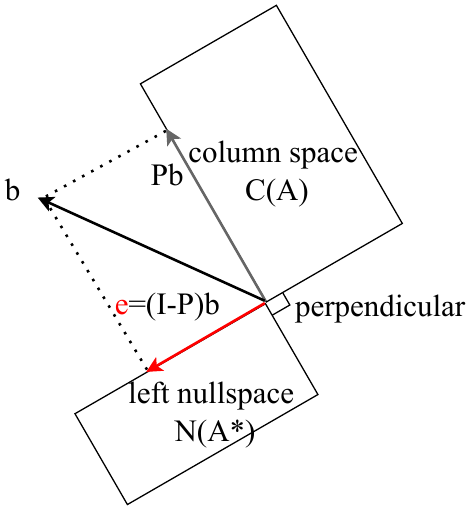}
  \caption{The Geometric Interpretation of Least Squares}
  \label{fig:SpaceVis}
\end{figure}
$\boldsymbol{E}_{\ell}\boldsymbol{z}_{\ell}$ can be considered as approximately equivalent to the residual of $\boldsymbol{z}_{\ell}$ projected onto the complement of the spanned space. 
$\boldsymbol{z}_{\ell} = \boldsymbol{z}_{\ell} + \eta \boldsymbol{E}_{\ell}\boldsymbol{z}_{\ell}$ represents the sample moving towards the complement of the span, thereby increasing the ranks of $\boldsymbol{Z}_{\ell}$ and $\{\boldsymbol{Z}_{\ell}\boldsymbol{\Pi}^j\}_{j=1}^{k}$, expanding the spanned space.
In contrast, $\boldsymbol{z}_{\ell} = \boldsymbol{z}_{\ell} - \eta \boldsymbol{C}_{\ell}^j\boldsymbol{z}_{\ell}$ shifts the sample towards the span of class  $j$, compressing the spanned subspace.
This can be verified by the equation below:
\begin{equation}
  \label{eq:EzExpandResiduals}
  \begin{split}
    \boldsymbol{E}_{\ell}\boldsymbol{z}_{\ell} 
    &= \alpha(\boldsymbol{z}_{\ell} - \boldsymbol{Z}_{\ell}[\beta_\ell]_*)\\
    &=\alpha(I-\alpha \boldsymbol{Z}_{\ell}(I+\alpha \boldsymbol{Z}_{\ell}^T\boldsymbol{Z}_{\ell})^{-1}\boldsymbol{Z}_{\ell}^T)\boldsymbol{z}_{\ell}
  \end{split}
\end{equation}
where $[\beta_\ell]_*= \alpha (I + \alpha \boldsymbol{Z}_{\ell}^T \boldsymbol{Z}_{\ell})^{-1}  \boldsymbol{Z}_{\ell}^T \boldsymbol{z}_{\ell}$.
Eq.\ref{eq:EzExpandResiduals} is exactly the \textit{residuals} of the ridge regression problem $[\beta_\ell]_* = \arg \min_{\beta_\ell} \alpha \|\boldsymbol{z}_\ell - \boldsymbol{Z}_\ell \beta_\ell\|_2^2 + \|\beta_\ell\|_2^2$.
Since the loss function of ridge regression is essentially a least squares function with an L2 norm regularization term, we can explain the geometric meaning of $\boldsymbol{E}_{\ell}\boldsymbol{z}_{\ell}$ by comparing it to the geometric interpretation of least squares.
As shown in Figure \ref{fig:SpaceVis}, for a classical least squares problem of $Ax = b$, the projection matrix $I-P= I-A(A^TA)^{-1}A^T$ projects vector $b$ onto the complement of column space $C(A)$, namely the left nullspace $N(A^T)$, and the projection is the error (or residual) $e = (I-P)b$ \cite{strang_linear_2012}. 
Thus, this confirms the geometric interpretation of $\boldsymbol{E}_{\ell}\boldsymbol{z}_{\ell}$.
For a detailed derivation, please see Appendix A.

As mentioned in the Background section, while the \textit{lifting} operation increases the upper limit of the dimensions of the spanned space, it is the expansion operator that truly enlarges the spanned space. 
Adequate expansion of the space is crucial to ensure the decoupling of various class subspaces and to ensure that samples are compressed into the correct subspaces. 
This highlights the greater importance of the expansion operator compared to the compression operator. 
Therefore, we introduce a weight function to dynamically control the expansion process.
\subsubsection{Weight Function for Controlling Expansion Dynamically}
A straightforward strategy 
is to enlarge the expansion when training faces stagnation.
However, this still results in samples undergoing incorrect transformations across multiple layers.
Therefore, it is preferable to minimize the misclassification rate of the estimation functions as quickly as possible.
When errors in membership estimation are present, we tentatively expand the space slowly. 
This helps maintain the compactness of samples within classes and facilitates effective ongoing training.
However, we also seek to rapidly reduce the misclassification rate of the estimatied function to prevent excessive updates in the wrong direction. 
Thus, we introduce a truncated, monotonically increasing exponential function as the weight function $w(\tau_{\ell})$ of the expansion operator $\boldsymbol{E}_{\ell}$, where
\begin{equation}
    w(\tau_{\ell}) = \min(\exp(\tau_{\ell}), u) \in [1, u], \tau_{\ell} \in [0, \infty).
    \label{eq:weightfunction}
\end{equation}
Here, $\tau_{\ell}$ increases with the number of layers, indicating that if the the subspaces do not decouple promptly, the weight of the expansion operator should be increased to facilitate more extensive expansion.
The product of $u$ and the learning rate $\eta$ should be less than or equal to 1 to avoid excessively large gradients. Specifically, when $\eta = 0.1$, $u$ can be set to 10.

It is important to emphasize that the gradient update $g(\boldsymbol{z}_{\ell}, \boldsymbol{\theta}_{\ell})$ of features is jointly determined by  $\boldsymbol{E}_{\ell}\boldsymbol{z}_{\ell}$ and $- \sum_{j=1}^{k}\gamma_j \boldsymbol{C}^{j}_{\ell}\boldsymbol{z}_{\ell}\boldsymbol{\pi}^{j} (\boldsymbol{z}_{\ell})$.
Due to their sum potentially being small or even zero, this can result in small or vanishing gradients. 
Therefore, increasing the weight of the expansion operator $\boldsymbol{E}_{\ell}$ helps enhance the magnitude of the gradient $g(\boldsymbol{z}_{\ell}, \boldsymbol{\theta}_{\ell})$ and expand the overall spanned space, improving the separability of subspaces and the accuracy of the estimation function. 
\subsection{Condition Number for Stopping training}
In addition to the objective function, the stability of network parameters can serve as a criterion for assessing network convergence.
A lower condition number indicates a more stable linear system and can serve as an auxiliary criterion to halt training.

For a ridge regression problem $[\beta]_* = \arg \min_{\beta} \left( \| y - X\beta \|_2^2 + \lambda \|\beta\|_2^2 \right)$, the condition number  is given by  $\kappa(X^T X + \lambda I) = \frac{\sigma_{\max}}{\sigma_{\min}}.$
Here, $\sigma_{\max}$ and $\sigma_{\min}$ are the maximum and minimum singular values of $X^T X + \lambda I$, respectively \cite{tabeart_improving_2019}. 
Due to the previously analyzed connection between ridge regression and network parameters, $k+1$ condition numbers of $(\boldsymbol{I}+\alpha\boldsymbol{Z}_{\ell}\boldsymbol{Z}^{T}_{\ell})$ and $\{(\boldsymbol{I}+\alpha_j\boldsymbol{Z}_{\ell}\mathbf{\Pi}^{j}\boldsymbol{Z}^{T}_{\ell})\}^k_{j=1}$ in the \( \boldsymbol{E}_{\ell} \) and $\{\boldsymbol{C}_{\ell}^j\}^k_{j=1}$ are measured to track the stability of the network. 
The condition number decreases as the number of layers increases. 
When there is no significant change in the condition numbers, it is considered that the network has converged, and training can be stopped. 
In practice, since calculating condition numbers is computationally intensive, it is feasible to assess the stability of network parameters every several tens of layers. 


\subsection{Algorithm Pseudocode}
\begin{algorithm}[htb]
  \caption{Training Phase}
  \label{alg:Train}
    \textbf{Input}: \(X = [x^1, \ldots, x^m] \in \mathbb{R}^{D  \times m}\), $\mathbf{\Pi}$, \(\epsilon > 0\), \(\lambda\), \(\eta\).\\
    \textbf{Output}: $\boldsymbol{Z}_{L+1}$, $\{\boldsymbol{E}_{\ell}\}^L_{\ell=1}$,$\{\boldsymbol{C}_{\ell}^j\}_{j=1,\ell=1}^{k,L}$,$\{w_{\ell}\}_{\ell=1}^{L}$,\\$\{p^{ij}_{\ell}\}_{i=1,j=1,\ell=1}^{k,k,L}$.
    \begin{algorithmic}[1]
     \STATE Set $\boldsymbol{Z}_1 \overset{\cdot}{=} [\boldsymbol{z}^1_1,\dots,\boldsymbol{z}_1^m] = \boldsymbol{X} \in \mathbb{R}^{n \times m}$, $\tau = 0$.
    \FOR{$\ell = 1,\dots,L$}
    \STATE  Compute parameters $\boldsymbol{E}_{\ell}$ and $\{\boldsymbol{C}_{\ell}^j\}^k_{j=1}$.
    \STATE  Assess the error rates of the estimation functions.
    \IF{ have errors}
        \STATE  Compute posterior probability 
        $\{p^{ij}_{\ell}\}_{i=1,j=1}^{k,k}$
        \FOR{$i = 1,\dots,m$}
        \STATE  Compute corrected estimations $\{p_c^j(\boldsymbol{z}_{\ell}^i)\}_{j=1}^{k}$.
        \STATE \# Controlling expansion dynamically
        \STATE $w_{\ell}(\tau_{\ell}) = \min(\exp(\tau_{\ell}), u)$
        \STATE \scriptsize $\boldsymbol{z}^i_{\ell+1} =
      \mathcal{P}_{\mathbb{S}^{n-1}}
      \left(
      \boldsymbol{z}^i_{\ell} + \eta \Big(w_{\ell} \boldsymbol{E}_{\ell}\boldsymbol{z}^i_{\ell} - \sum_{j=1}^{k}\gamma_j \boldsymbol{C}^{j}_{\ell}\boldsymbol{z}^i_{\ell}p_c^j(\boldsymbol{z}_{\ell}^i) \Big)
      \right)
      ;$
      \normalsize
      \ENDFOR
      \STATE $\tau_{\ell+1}=\tau_{\ell}+0.1$
    \ELSIF{ do not have errors}
        \FOR{$i = 1,\dots,m$}
       \STATE \scriptsize $\boldsymbol{z}^i_{\ell+1} =
      \mathcal{P}_{\mathbb{S}^{n-1}}
      \left(
      \boldsymbol{z}^i_{\ell} + \eta \Big(\boldsymbol{E}_{\ell}\boldsymbol{z}^i_{\ell} - \sum_{j=1}^{k}\gamma_j \boldsymbol{C}^{j}_{\ell}\boldsymbol{z}^i_{\ell}\hat{\boldsymbol{\pi}}^{j} (\boldsymbol{z}^i_{\ell}) 
      \Big)
      \right)
      ;$
      \normalsize
        \ENDFOR
    \ENDIF
    \IF{no change on condition number}
    \STATE  stop training
    \ENDIF
    \ENDFOR
  \end{algorithmic}
\end{algorithm}
As the Algorithm \ref{alg:Train} shows,
the errors of the estimation functions are calculated (line 4).
If there are misclassified samples, incorporation of label knowledge and the dynamic control of expansion are employed (lines 5-13).
Otherwise, the original method is executed (lines 14-18). 
Condition number serves as an auxiliary criterion for stopping training (lines 19-21) and  does not need to computed at every layer.
Moreover, Bayesian inference and dynamic expansion only involve basic arithmetic operations.
Therefore, ESS-ReduNet does not significantly increase the computational load.
The detailed condition number curves presented in the experimental section are intended solely to highlight the acceleration effects of ESS-ReduNet.

\section{Experiments}
\label{sec:experiments}
ESS-ReduNet is evaluated on seven datasets: HAR \cite{jorge_reyes-ortiz_human_2012}, ESR \cite{qiuyi_wu_epileptic_2017}, Covertype (C) \cite{blackard_covertype_1998}, Gas \cite{vergara_gas_2012}, mfeatFactors and mfeatFourier \cite{duin_multiple_1998},
musk \cite{david_chapman_musk_1994}. 
We use \(\epsilon^2 = 0.1\) and \(\eta = 0.1\) for all datasets.
The initial number of layers is set to 3000,
which is sufficient to highlight the strengths of our method.

A case study on the ESR dataset is used to demonstrate how our method accelerates network training and produces higher-quality features.
In addition, an ablation study is conducted to underscore the necessity of Bayesian inference and dynamic control of the expansion.
Finally, based on \cite{yu_learning_2020}, we detail the classification accuracy of Support Vector Machine (SVM), K-Nearest Neighbors (KNN), and Nearest Subspace Classifier (NSC) across seven datasets, demonstrating the acceleration and corrective benefits of ESS-ReduNet.

\subsection{Case Study}
\begin{figure}[htb]
  \centering
  \begin{subfigure}{0.49\linewidth}
  \centering
    \includegraphics[width=\linewidth]{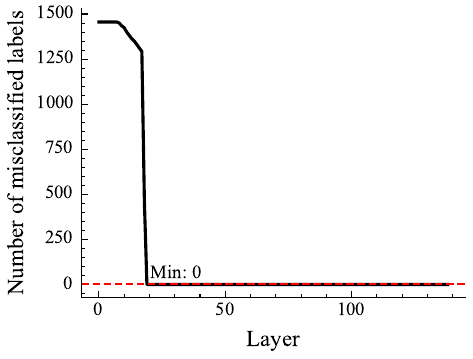}
    \caption{}
    \label{fig:BayesExpandofYerr}
  \end{subfigure}
  \begin{subfigure}{0.49\linewidth}
  \centering
    \includegraphics[width=\linewidth]{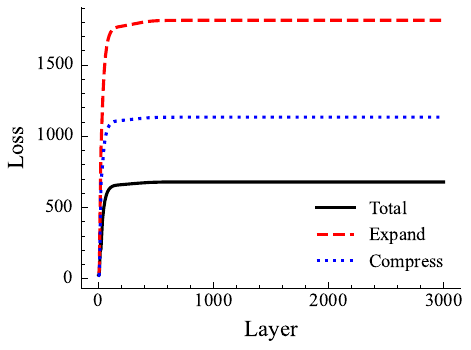}
    \caption{}
    \label{fig:MCRofBayesExpandESR}
  \end{subfigure}
  
  \begin{subfigure}{0.49\linewidth}
  \centering
    \includegraphics[width=\linewidth]{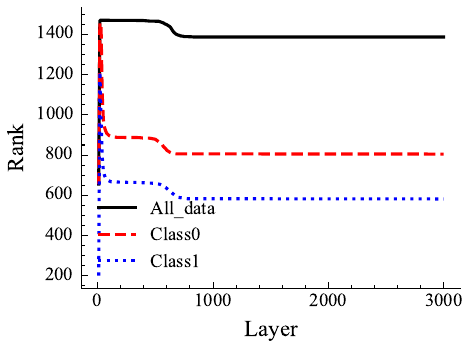}
    \caption{}
    \label{fig:RankofBayesExpandESR}
  \end{subfigure}
  \begin{subfigure}{0.49\linewidth}
  \centering
    \includegraphics[width=\linewidth]{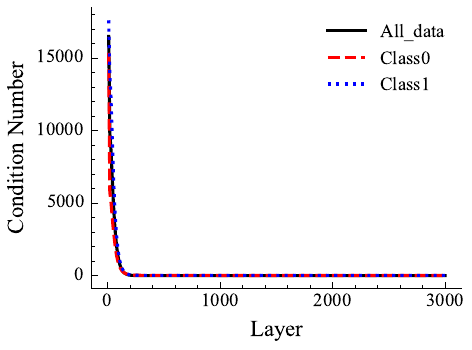}
    \caption{}
    \label{fig:ConditionNumberofBayesExpandESR}
  \end{subfigure}
  \caption{Performance of ESS-ReduNet on ESR Dataset.
  }
  \label{fig:BayesExpand}
\end{figure}
\begin{figure}[hbt]
  \centering
  \begin{subfigure}{0.4\linewidth}
  \centering
    \includegraphics[width=\linewidth]{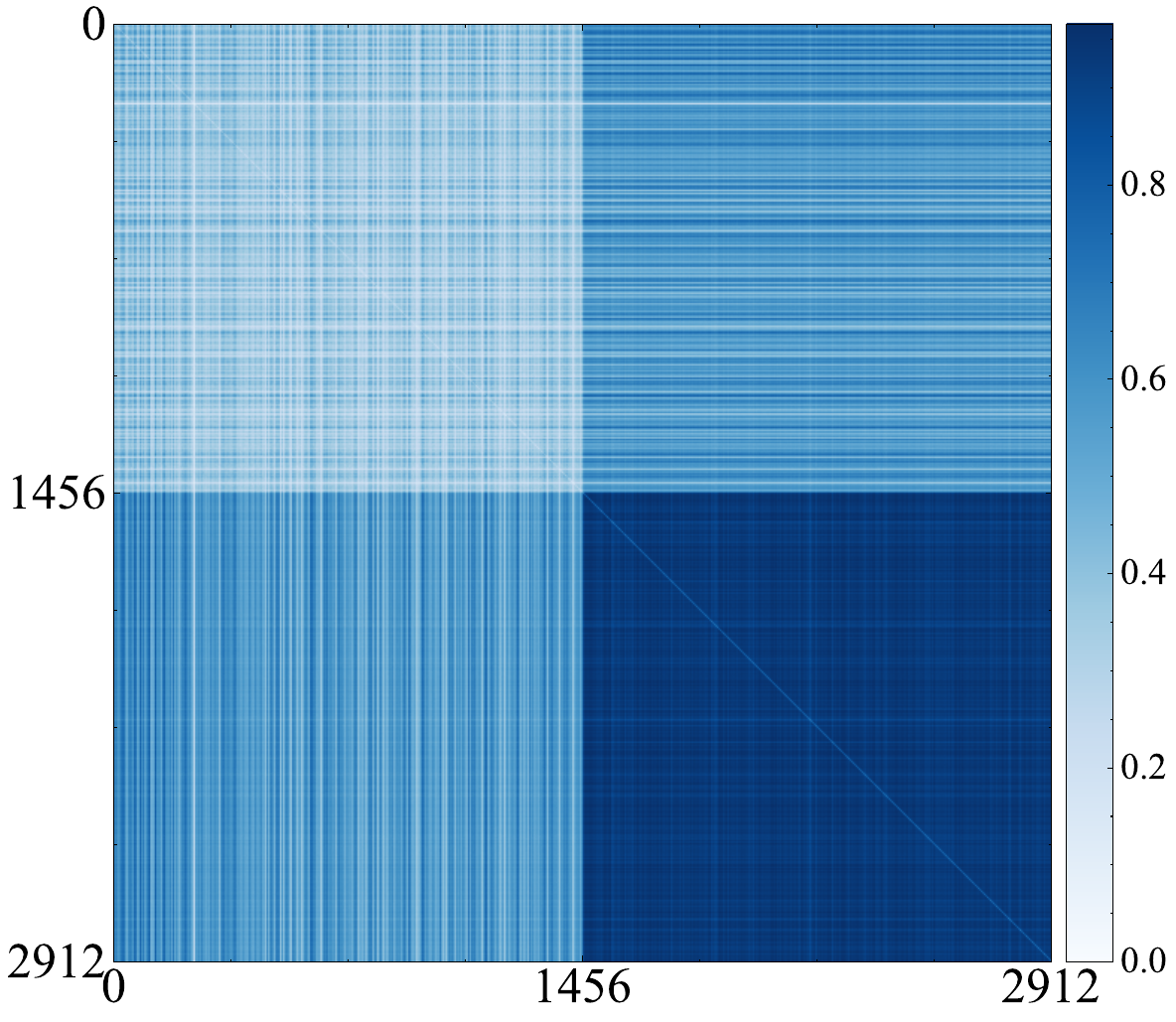}
    \caption{ReduNet}
    \label{fig:heatmapofOrigin}
  \end{subfigure}
  \begin{subfigure}{0.4\linewidth}
  \centering
    \includegraphics[width=\linewidth]{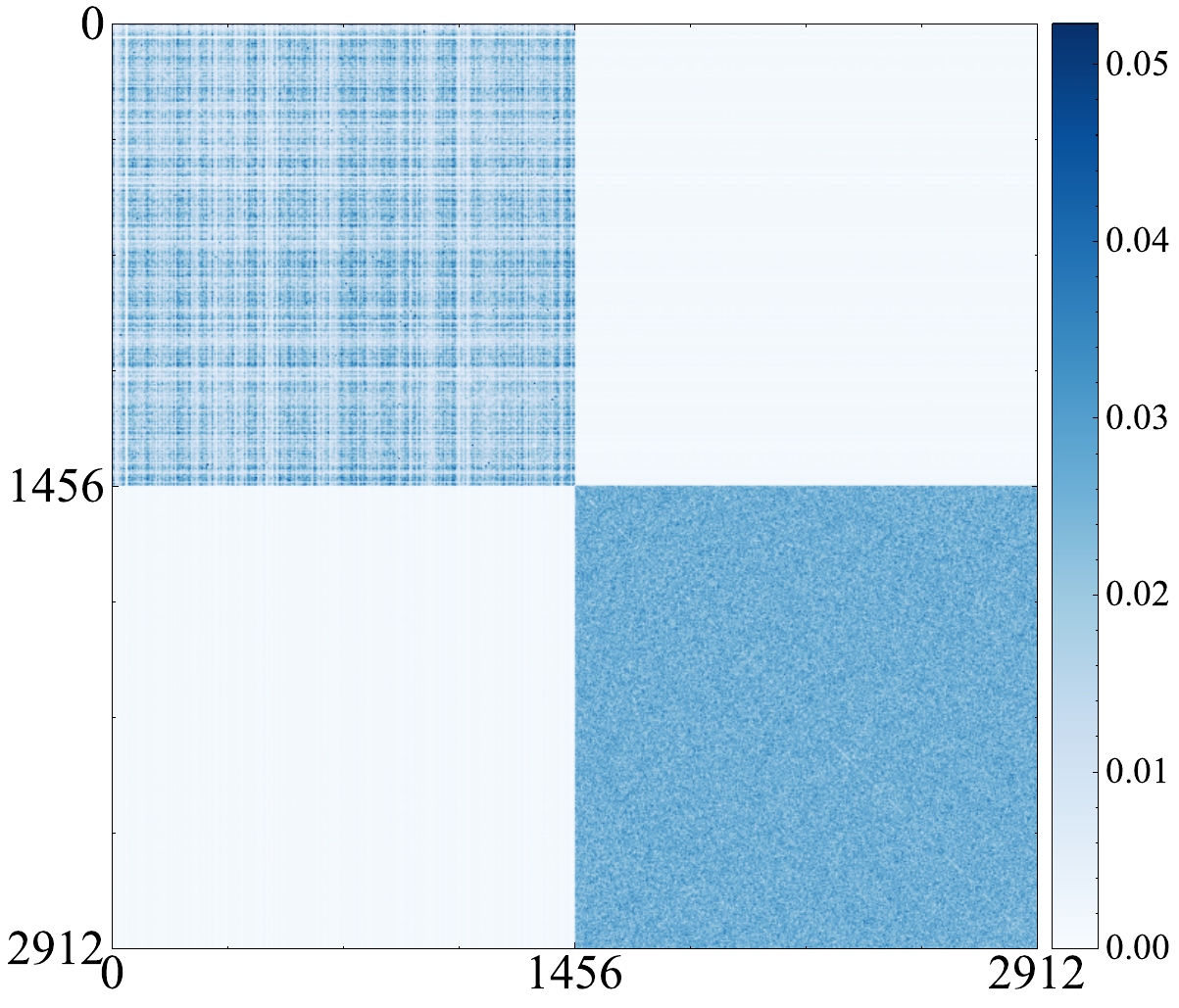}
    \caption{ESS-ReduNet}
    \label{fig:heatmapofBayesExpand}
  \end{subfigure}
  \caption{Visualization of ESR Feature Orthogonalization at the 200th Layer. Heatmap of $|\boldsymbol{Z} \boldsymbol{Z}^*|$ showing correlation levels: darker shades indicate higher correlation between two samples. Ideally, the subspaces of two classes should be orthogonal,
    evident as a block diagonal structure.
  }
  \label{fig:heatmaps}
\end{figure}
Same with the Figure \ref{fig:test}, four aspects are evaluated on ESS-ReduNet. 
Notably, 
the number of misclassified samples rapidly decreases, reaching zero by the 19th layer as shown in Figure \ref{fig:BayesExpandofYerr}.
In comparison, ReduNet required 2000 layers to reduce the misclassification count to 495, as demonstrated in Figure \ref{fig:originofYerr}, highlighting the accelerated network convergence achieved with ESS-ReduNet.
Besides, as illustrated in Figure \ref{fig:MCRofBayesExpandESR}, our method also achieves a higher $\mathrm{MCR}^2$ value.

Figure \ref{fig:RankofBayesExpandESR} confirms that ESS-ReduNet effectively increases the rank of the linear space and expands the spanned space.
The dimensions of the spanned spaces for the two classes add up to the total space dimension. 
As evident in Figure \ref{fig:heatmapofBayesExpand}, the two subspaces are now complementary and occupy the entire space.
This suggests that, according to Figure \ref{fig:ConditionNumberofBayesExpandESR}, it is reasonable to halt network training when there is no significant change in the condition number, which occurs around the 200th layer.
\begin{figure}[htb]
  \centering
  \begin{subfigure}{0.49\linewidth}
  \centering
    \includegraphics[width=\linewidth]{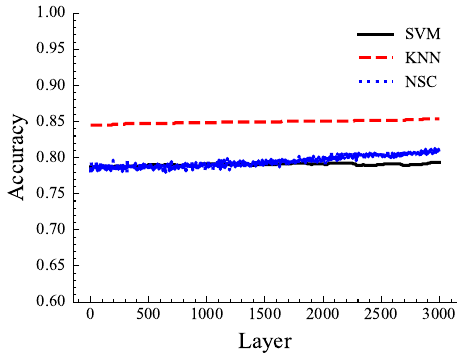}
    \caption{ReduNet}
    \label{fig:accFourierOrigin}
  \end{subfigure}
  \begin{subfigure}{0.49\linewidth}
  \centering
    \includegraphics[width=\linewidth]{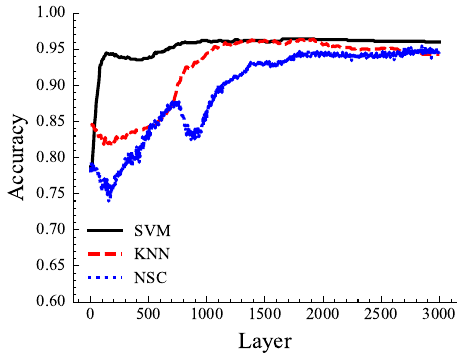}
    \caption{ESS-ReduNet}
    \label{fig:accFourierBayesExpand}
  \end{subfigure}
  \caption{Performance Comparison on the Fourier Version. 
  }
  \label{fig:performesofFouriers}
\end{figure}

In fact, our improvements is a plug-in approach that can be conveniently combined with the Fourier version of ReduNet. 
Although the convergence rate of the Fourier version is significantly slower, 
we have tested our method on the Fourier version as well. 
By comparing Figures \ref{fig:accFourierOrigin} and \ref{fig:accFourierBayesExpand}, it is evident that our method achieves higher accuracy.
\subsection{Ablation Study}
\begin{figure}[htb]
  \centering
  \begin{subfigure}{0.49\linewidth}
  \centering
    \includegraphics[width=\linewidth]{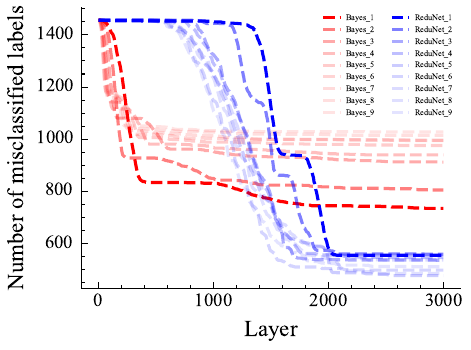}
    \caption{}
    \label{fig:YerrBayesOrigin}
  \end{subfigure}
  \begin{subfigure}{0.49\linewidth}
  \centering
    \includegraphics[width=\linewidth]{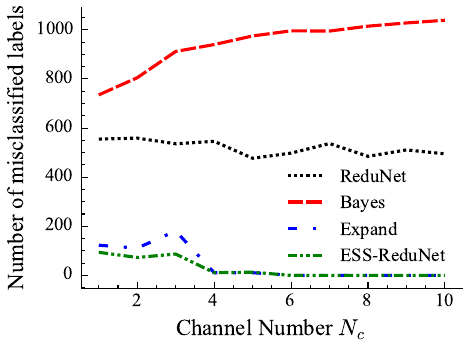}
    \caption{}
    \label{fig:LowestYerr}
  \end{subfigure}
  \caption{(a):The misclassification number of estimation functions for both the \textit{Bayes} method (red line) and ReduNet (blue line). Different opacities represent varying channel numbers $N_c$ of \textit{lifting} operation; 
  As $N_c$ increases, the color becomes lighter.
  (b):The lowest number of misclassified samples by estimation functions for four methods, under different channel numbers $N_c$.
  }
  \label{fig:AblationStudy}
\end{figure}
The following combinatorial approaches are evaluated to show the necessity of the Bayesian inference and dynamic expansion modules: ReduNet with Bayesian inference (\textit{Bayes}), ReduNet with expansion dynamically (\textit{Expand}), ESS-ReduNet and ReduNet.
As shown by the red line in Figure \ref{fig:YerrBayesOrigin}, the \textit{Bayes} method accelerates the reduction of misclassified samples compared to ReduNet, but subsequently reaches a plateau, exhibiting a phenomenon similar to gradient vanishing.
Nevertheless,
as displayed in Figure \ref{fig:LowestYerr}, 
the ESS-ReduNet
performs better under conditions of lower channel numbers than \textit{Expand} method.
This indicates that enlarged spanned space at higher channel numbers enhances subspaces' separability. 
In contrast, under low channel conditions, the label knowledge from \textit{Bayes} method demonstrates its effectiveness in a limited spanned space.


Furthermore, Figure \ref{fig:YerrBayesOrigin} shows that as the channel number increases, both the \textit{Bayes} method and the ReduNet demonstrate a faster decline in the estimation function's error, as evidenced by the leftward shift of the red and blue curves. 
This observation confirms our theoretical analysis, suggesting that a higher number of channels, indicative of more convolutional kernels, provides additional bases that enhance the separation of samples across different classes.

\subsection{Performance on Classification Tasks}

\begin{table}[hbt]
\setlength{\tabcolsep}{1mm}
\centering
\begin{tabular}{lrcccr}
\toprule
Dataset      & {\(\mathrm{MCR}^2\)}             & SVM           & KNN           & NSC           & Layer        \\
\midrule
ESR         & 245.89           & 0.65          & 0.63          & 0.72          & $>3000$         \\
             & \textbf{660.84}  & \textbf{0.96} & \textbf{0.78} & \textbf{0.78} & \textbf{199} \\
HAR          & 989.27           & 0.68          & 0.75          & 0.73          & $>3000$         \\
             & \textbf{1066.02} & \textbf{0.70} & \textbf{0.75} & \textbf{0.75} & \textbf{326} \\
Covertype    & 108.09           & 0.54          & 0.71          & 0.56          & $>3000$        \\
             & \textbf{295.79}  & \textbf{0.74} & \textbf{0.74} & \textbf{0.71} & \textbf{219} \\
Gas          & 147.12           & \textbf{0.98} & \textbf{0.99} & 0.95          & $>3000$         \\
             & \textbf{889.90}  & 0.97          & 0.97          & \textbf{0.96} & \textbf{94}  \\
mfeatFactors & 603.38           & 0.97          & 0.97          & 0.97          & 194          \\
             & \textbf{1011.57} & \textbf{0.97} & \textbf{0.97} & \textbf{0.97} & \textbf{125} \\
mfeatFourier & 668.10           & 0.82          & 0.82          & 0.82          & 134          \\
             & \textbf{727.35}  & \textbf{0.82} & \textbf{0.82} & \textbf{0.82} & \textbf{134} \\
musk         & 481.99           & \textbf{0.93} & 0.93          & 0.93          & 164          \\
             & \textbf{624.44}  & 0.91          & \textbf{0.93} & \textbf{0.93} & \textbf{119} \\
             \bottomrule
\end{tabular}

\caption{Comparative results on \(\mathrm{MCR}^2\), accuracy on three classifiers (higher value is better) and the layer of stopping training (lower value is better), with the first row showing outcomes from ReduNet and the second row presenting results of ESS-ReduNet.}
\label{tab:ACCalldata}
\end{table}
Table \ref{tab:ACCalldata}  displays the test results of ESS-ReduNet across seven datasets. 
The \textit{Layer} column indicates the convergence in the condition number. 
As shown in the table, ESS-ReduNet achieves higher \(\mathrm{MCR}^2\) values across all datasets.
Besides, it allows training to stop earlier than ReduNet. Finally, ESS-ReduNet has achieved varying degrees of improvement in accuracy on three basic classifiers, or achieves acceleration while maintaining comparable accuracy to ReduNet.

\section{Conclusion}
\label{sec:conclusion}
This paper addresses issues arising from inaccurate estimation functions in ReduNet, such as incorrect feature updates, slow training, and the resultant poor quality of network structures and transformed features. 
The performance of the estimation functions depends on the separability of the subspaces of various classes.
Hence, we propose the ESS-ReduNet: a improved framework aims to enhance the subspace separability via controlling the expansion dynamically and incorporating the label knowledge through Bayesian inference. 
Both encourage the decoupling of subspaces, thereby improving the estimation performance.
Finally, we track changes in the condition number of network parameters as a criterion to halt network training.
As a plug-in approach, our improvements can be easily integrated with the Fourier version of ReduNet.
Experimental results show that our method significantly accelerates network training,  improves feature quality, and conserves computational resources.

\bibliographystyle{named}
\bibliography{references}

\appendix
\section{Detailed Derivation of the Geometric Interpretation of Expansion Operators}
\begin{figure}[hbt]
  \centering
    \includegraphics[width=0.45\linewidth]{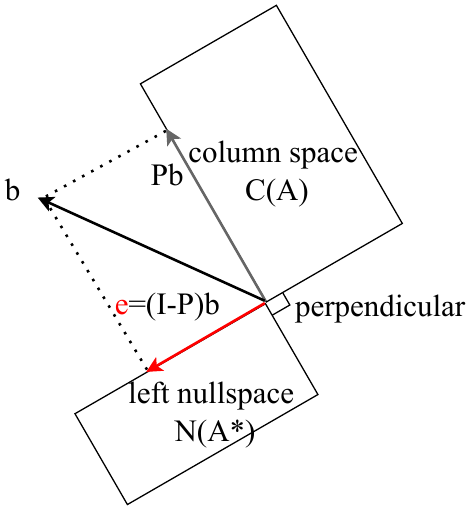}
  \caption{The Geometric Interpretation of Least Squares}
  \label{fig:SpaceVis2}
\end{figure}
Although \citeauthor{chan_redunet_2022} \shortcite{chan_redunet_2022} have discussed the relationship between expansion operators and ridge regression, we further clarify their geometric interpretation by comparing the forms of the expansion operators with those of least squares and ridge regression.

For a least squares problem of $Ax = b$, the projection matrix is $P=A(A^*A)^{-1}A^*$. As show in Figure \ref{fig:SpaceVis2}, its geometric meaning is to project vector $b$ onto the column space $C(A)$. The matrix $I-P$ is also a projection matrix, which projects vector $b$ onto the complement of column space $C(A)$, namely the left nullspace $N(A^*)$, and the projection is the error (or residual) $e = (I-P)b$ \cite{strang_linear_2012}. 
For a feature  $\boldsymbol{z}_{\ell}$,
$\boldsymbol{E}_{\ell}\boldsymbol{z}_{\ell}$ can be expanded using the Woodbury identity \cite{van_wieringen_lecture_2023} $(I + UV)^{-1} = I - U (I + VU)^{-1} V$ as follows:
\begin{equation}
  \label{eq:EzExpand}
  \begin{split}
    \boldsymbol{E}_{\ell}\boldsymbol{z}_{\ell} 
    &=\alpha(I-\alpha \boldsymbol{Z}_{\ell}(I+\alpha \boldsymbol{Z}_{\ell}^*\boldsymbol{Z}_{\ell})^{-1}\boldsymbol{Z}_{\ell}^*)\boldsymbol{z}_{\ell}.
  \end{split}
\end{equation}
This is exactly the \textit{residuals} of the ridge regression problem $[\beta_\ell]_* = \arg \min_{\beta_\ell} \alpha \|\boldsymbol{z}_\ell - \boldsymbol{Z}_\ell \beta_\ell\|_2^2 + \|\beta_\ell\|_2^2$, which is
\begin{equation}
  \label{eq:ridgeError}
  \small
  \begin{split}
    \alpha(\boldsymbol{z}_{\ell} - \boldsymbol{Z}_{\ell}[\beta_\ell]_*)
    &=\alpha(I- \alpha \boldsymbol{Z}_{\ell}(I+ \alpha \boldsymbol{Z}_{\ell}^*\boldsymbol{Z}_{\ell})^{-1}\boldsymbol{Z}_{\ell}^*)\boldsymbol{z}_{\ell}
  \end{split}
  \normalsize
\end{equation}
\normalsize
where $[\beta_\ell]_*= \alpha (I + \alpha \boldsymbol{Z}_{\ell}^* \boldsymbol{Z}_{\ell})^{-1}  \boldsymbol{Z}_{\ell}^* \boldsymbol{z}_{\ell}.$ 
Since the loss function of ridge regression is essentially a least squares function with an L2 norm regularization term, by analogy with the residual projection matrix $(I-P)$, we can conclude that $\boldsymbol{E}_{\ell} \boldsymbol{z}_{\ell}$ approximates the projection of $\boldsymbol{z}_{\ell}$'s residual onto the complement of the spannd space.

\subsection{The Solution of Ridge Regression}
The objective function of ridge regression problem  in Eq.\ref{eq:ridgeError} is given by:

$$
[\beta_\ell]_* = \arg \min_{\beta_\ell} \alpha \|\boldsymbol{z}_\ell - \boldsymbol{Z}_\ell \beta_\ell\|_2^2 + \|\beta_\ell\|_2^2
$$
where $\boldsymbol{Z}_{\ell}$ represents the design matrix encompassing all data, $\boldsymbol{z}_{\ell}$ denotes the response vector (which, in this paper, corresponds to a feature), and $\alpha$ is the regularization parameter.
The partial derivative with respect to $\beta_{\ell}$ is:
$$
-2 \alpha \boldsymbol{Z}_{\ell}^* (\boldsymbol{z}_{\ell} - \boldsymbol{Z}_{\ell} \beta_{\ell}) + 2 \beta_{\ell}
$$
Rearrange the gradient to zero to solve for $\beta_{\ell}$:
$$
-2 \alpha \boldsymbol{Z}_{\ell}^* \boldsymbol{z}_{\ell} + 2 \alpha \boldsymbol{Z}_{\ell}^* \boldsymbol{Z}_{\ell} \beta_{\ell} + 2 \beta_{\ell} = 0
$$
Rearranging terms to isolate $\beta_{\ell}$ on one side yields:
$$
(\alpha \boldsymbol{Z}_{\ell}^* \boldsymbol{Z}_{\ell} + I) \beta_{\ell} = \alpha \boldsymbol{Z}_{\ell}^* \boldsymbol{z}_{\ell}
$$
Solving this equation, we have:
$$
[\beta_\ell]_*= \alpha (I + \alpha \boldsymbol{Z}_{\ell}^* \boldsymbol{Z}_{\ell})^{-1}  \boldsymbol{Z}_{\ell}^* \boldsymbol{z}_{\ell}
$$
\section{The Algorithm of ESS-ReduNet}
The algorithm \ref{alg:test} is the testing phase of ESS-ReduNet.
Dynamic control is applied based on whether parameters related to Bayesian inference are detected.
\begin{algorithm}[htb]
  \caption{Testing Phase}
  \label{alg:test}
    \textbf{Input}: $\boldsymbol{x} \in \mathbb{R}^D$, network parameters $\{\boldsymbol{E}_{\ell}\}^L_{\ell=1}$, $\{\boldsymbol{C}_{\ell}^j\}_{j=1,\ell=1}^{k,L}$, $\{w_{\ell}\}_{\ell=1}^{L}$ and $\{p^{ij}_{\ell}\}_{i=1,j=1,\ell=1}^{k,k,L}$, feature dimension \(n\), \(\lambda\), and a learning rate \(\eta\).\\
    \textbf{Output}: features $\boldsymbol{z}_{L+1}$.
    \begin{algorithmic}[1]
    \STATE Set $\boldsymbol{z}_1 = \boldsymbol{x} \in \mathbb{R}^n$
    \FOR{$\ell = 1,\dots,L$}
    \IF{ need Bayes}
        \STATE  Compute corrected estimations $\{p_c^j(\boldsymbol{z}_{\ell})\}_{j=1}^{k}$.
        \STATE \scriptsize $\boldsymbol{z}_{\ell+1} =
      \mathcal{P}_{\mathbb{S}^{n-1}}
      \left(
      \boldsymbol{z}_{\ell} + \eta \Big(w_{\ell} \boldsymbol{E}_{\ell}\boldsymbol{z}_{\ell} - \sum_{j=1}^{k}\gamma_j \boldsymbol{C}^{j}_{\ell}\boldsymbol{z}_{\ell}p_c^j(\boldsymbol{z}_{\ell}) \Big)
      \right)
      ;$
      \normalsize
    \ELSIF{do not need Bayes}
       \STATE \scriptsize $\boldsymbol{z}_{\ell+1} =
      \mathcal{P}_{\mathbb{S}^{n-1}}
      \left(
      \boldsymbol{z}_{\ell} + \eta \Big(\boldsymbol{E}_{\ell}\boldsymbol{z}_{\ell} - \sum_{j=1}^{k}\gamma_j \boldsymbol{C}^{j}_{\ell}\boldsymbol{z}_{\ell}\hat{\boldsymbol{\pi}}^{j} (\boldsymbol{z}_{\ell}) 
      \Big)
      \right)
      ;$
      \normalsize
    \ENDIF
    \ENDFOR
  \end{algorithmic}
\end{algorithm}

\section{Introduction of Nearest Subspace Classifier}
ReduNet acts as a feature transformation function  \(f_{\theta}\). As demonstrated in \cite{yu_learning_2020},  each class’s representations occupy low-dimensional, mutually orthogonal subspaces. 
Therefore, assuming that the learned features satisfy the theoretical properties, for a test sample \(z_{test} = f_{\theta}(x_{test})\), we can classify the sample using the NSC. Formally, the predicted label is given by:
\begin{equation}
  y = \arg\min_{j \in \{1, \ldots, k\}} \left\| \left(I - U_jU_j^*\right)z_{test} \right\|_2^2
\end{equation}
Here, \(U_j \in \mathbb{R}^{n\times r_j}\) represents the first \(r_j\) principal components of learned feature \(\boldsymbol{Z_j}\) that corresponds to class \( j\).
\section{Detailed Experiment Results}
\subsection{Number of Misclassified Labels}
\begin{figure}[htb]
  \centering
  \begin{subfigure}{0.49\linewidth}
  \centering
    \includegraphics[width=\linewidth]{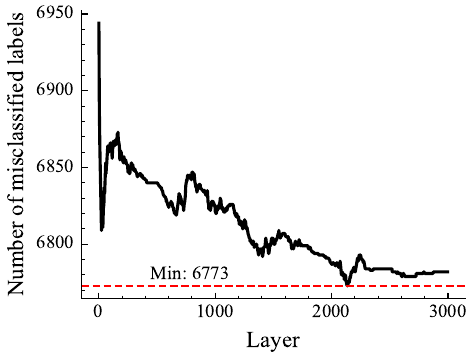}
    \caption{ReduNet}
    \label{fig:covtypepi}
  \end{subfigure}
  \begin{subfigure}{0.49\linewidth}
  \centering
    \includegraphics[width=\linewidth]{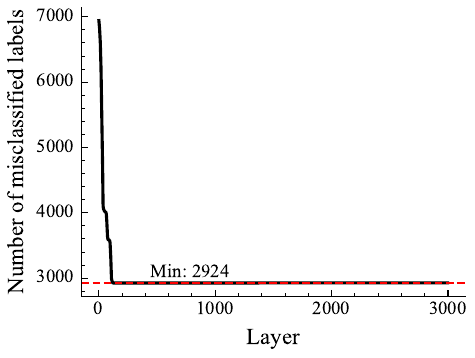}
    \caption{ESS-ReduNet}
    \label{fig:covtypepi2}
  \end{subfigure}
  \caption{covtype}
  \label{fig:picovtype}
\end{figure}

\begin{figure}[htb]
    \begin{subfigure}{0.49\linewidth}
      \centering
        \includegraphics[width=\linewidth]{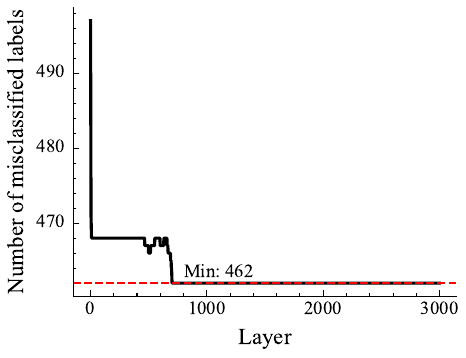}
        \caption{ReduNet}
        \label{fig:gaspi}
    \end{subfigure}
    \begin{subfigure}{0.49\linewidth}
      \centering
        \includegraphics[width=\linewidth]{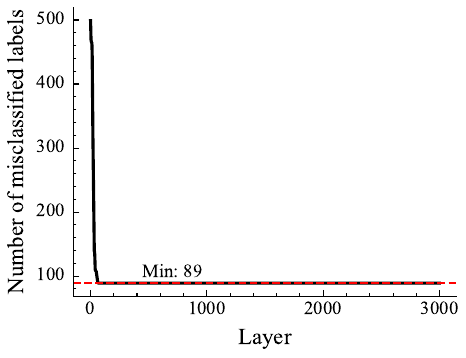}
        \caption{ESS-ReduNet}
        \label{fig:gaspi2}
    \end{subfigure}
  \caption{gas}
\end{figure}

\begin{figure}[htb]
    \begin{subfigure}{0.49\linewidth}
      \centering
        \includegraphics[width=\linewidth]{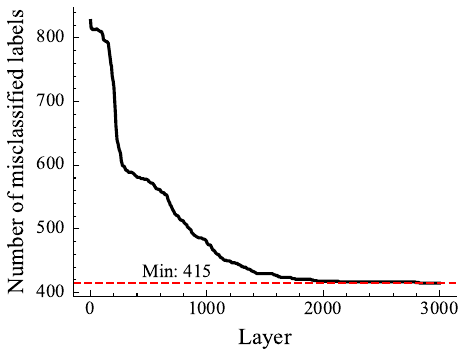}
        \caption{ReduNet}
        \label{fig:HARpi}
    \end{subfigure}
    \begin{subfigure}{0.49\linewidth}
      \centering
        \includegraphics[width=\linewidth]{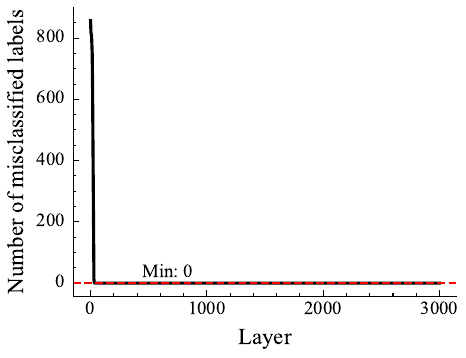}
        \caption{ESS-ReduNet}
        \label{fig:HARpi2}
    \end{subfigure}
  \caption{HAR}
\end{figure}

\begin{figure}[htb]
    \begin{subfigure}{0.49\linewidth}
      \centering
        \includegraphics[width=\linewidth]{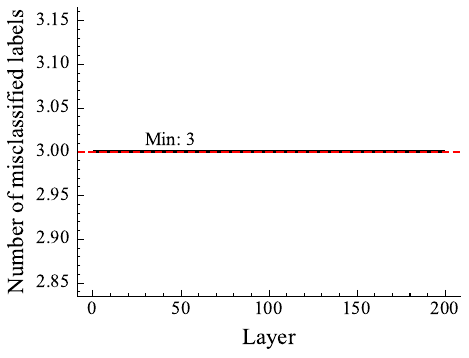}
        \caption{ReduNet}
        \label{fig:mfeatFactorspi}
    \end{subfigure}
    \begin{subfigure}{0.49\linewidth}
      \centering
        \includegraphics[width=\linewidth]{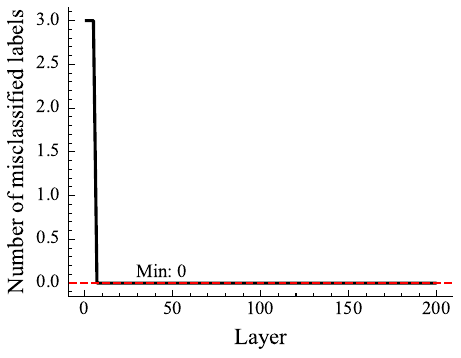}
        \caption{ESS-ReduNet}
        \label{fig:mfeatFactorspi2}
    \end{subfigure}
  \caption{mfeatFactors}
\end{figure}

\begin{figure}[htb]
    \begin{subfigure}{0.49\linewidth}
      \centering
        \includegraphics[width=\linewidth]{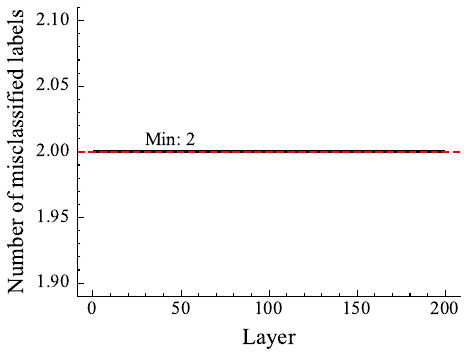}
        \caption{ReduNet}
        \label{fig:mfeatFourierspi}
    \end{subfigure}
    \begin{subfigure}{0.49\linewidth}
      \centering
        \includegraphics[width=\linewidth]{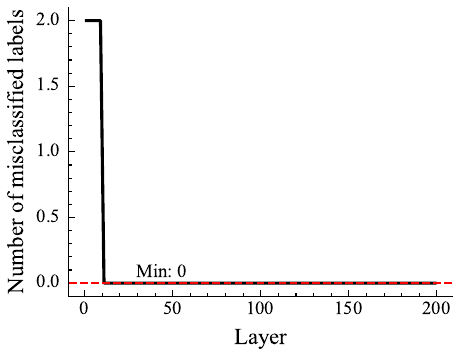}
        \caption{ESS-ReduNet}
        \label{fig:mfeatFourierpi2}
    \end{subfigure}
  \caption{mfeatFourier}
\end{figure}

\begin{figure}[htb]
    \begin{subfigure}{0.49\linewidth}
      \centering
        \includegraphics[width=\linewidth]{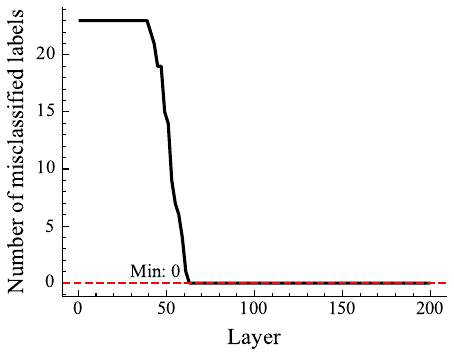}
        \caption{ReduNet}
        \label{fig:muskspi}
    \end{subfigure}
    \begin{subfigure}{0.49\linewidth}
      \centering
        \includegraphics[width=\linewidth]{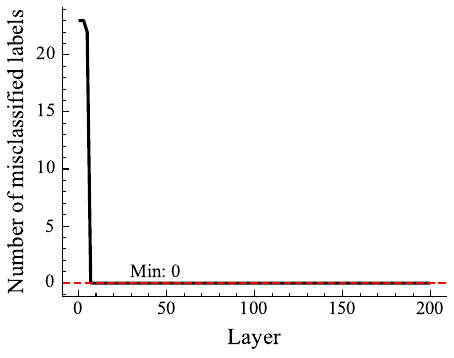}
        \caption{ESS-ReduNet}
        \label{fig:muskpi2}
    \end{subfigure}
  \caption{musk}
  \label{fig:pimusk}
\end{figure}

Figures \ref{fig:picovtype} to \ref{fig:pimusk} illustrate the decline in misclassification rates of the estimation functions across six datasets. It is evident that in ESS-ReduNet, the decrease in misclassification rates is faster, and it achieves lower misclassification rates compared to ReduNet.

\subsection{Objective Function Curve}
\begin{figure}[htb]
    \begin{subfigure}{0.49\linewidth}
      \centering
        \includegraphics[width=\linewidth]{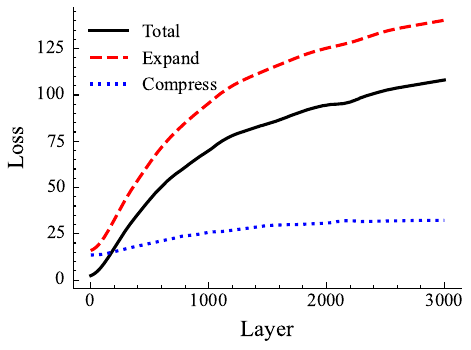}
        \caption{ReduNet}
        \label{fig:covtypemcr}
    \end{subfigure}
    \begin{subfigure}{0.49\linewidth}
      \centering
        \includegraphics[width=\linewidth]{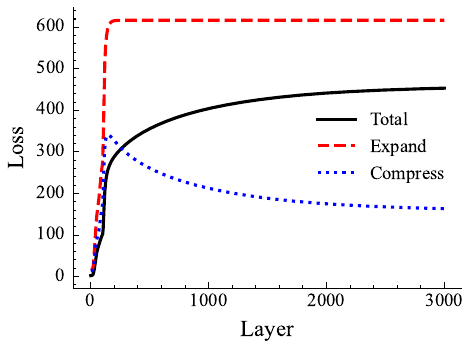}
        \caption{ESS-ReduNet}
        \label{fig:covtypemcr2}
    \end{subfigure}
  \caption{covtype}
  \label{fig:mcrcovtype}
\end{figure}

\begin{figure}[htb]
    \begin{subfigure}{0.49\linewidth}
      \centering
        \includegraphics[width=\linewidth]{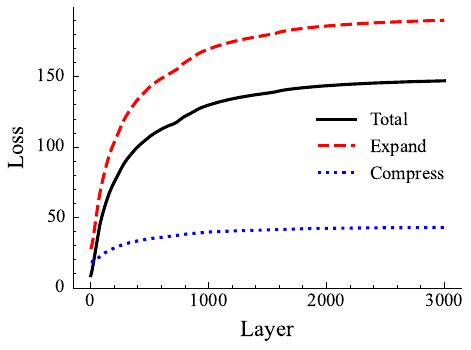}
        \caption{ReduNet}
        \label{fig:gasmcr}
    \end{subfigure}
    \begin{subfigure}{0.49\linewidth}
      \centering
        \includegraphics[width=\linewidth]{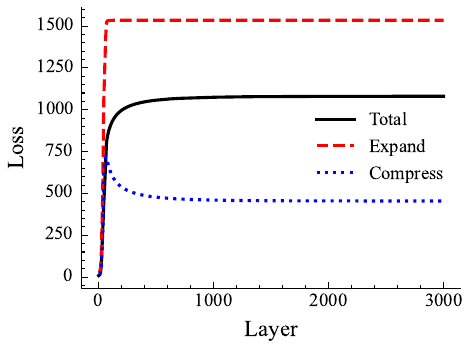}
        \caption{ESS-ReduNet}
        \label{fig:gasmcr2}
    \end{subfigure}
  \caption{gas}
\end{figure}

\begin{figure}[htb]
    \begin{subfigure}{0.49\linewidth}
      \centering
        \includegraphics[width=\linewidth]{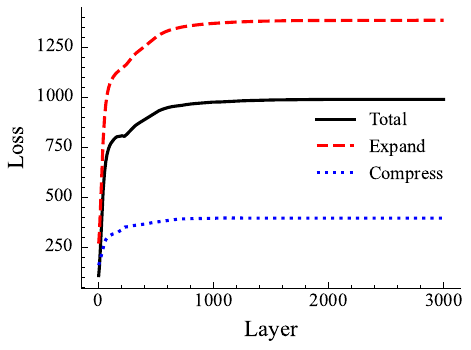}
        \caption{ReduNet}
        \label{fig:HARv1mcr}
    \end{subfigure}
    \begin{subfigure}{0.49\linewidth}
      \centering
        \includegraphics[width=\linewidth]{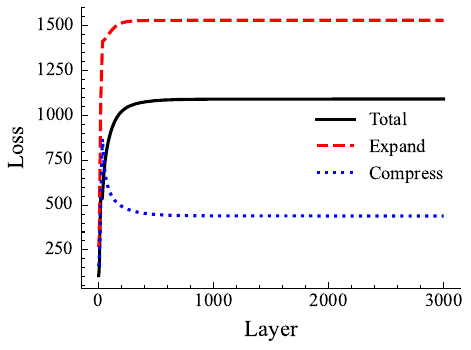}
        \caption{ESS-ReduNet}
        \label{fig:HARv1mcr2}
    \end{subfigure}
  \caption{HARv1}
\end{figure}

\begin{figure}[htb]
    \begin{subfigure}{0.49\linewidth}
      \centering
        \includegraphics[width=\linewidth]{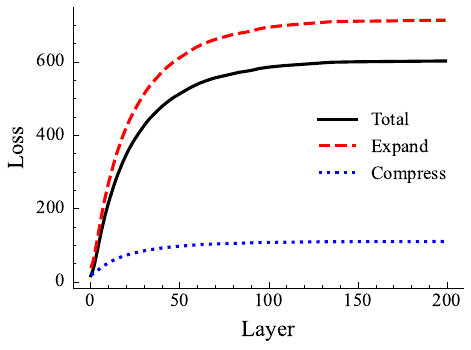}
        \caption{ReduNet}
        \label{fig:mfeatFactorsmcr}
    \end{subfigure}
    \begin{subfigure}{0.49\linewidth}
      \centering
        \includegraphics[width=\linewidth]{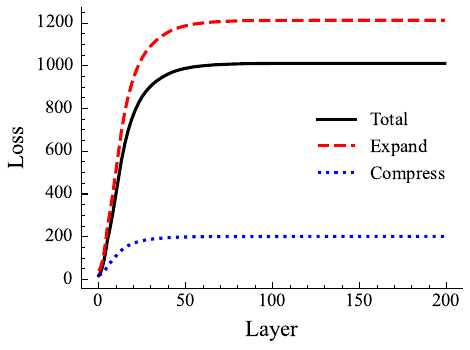}
        \caption{ESS-ReduNet}
        \label{fig:mfeatFactorsmcr2}
    \end{subfigure}
  \caption{mfeatFactors}
\end{figure}

\begin{figure}[htb]
    \begin{subfigure}{0.49\linewidth}
      \centering
        \includegraphics[width=\linewidth]{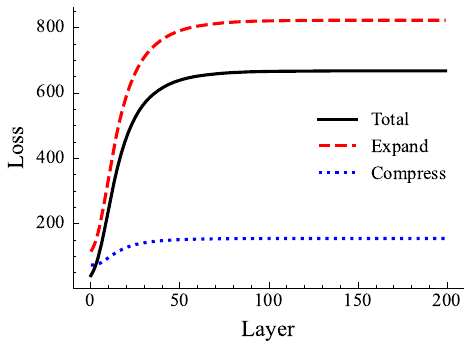}
        \caption{ReduNet}
        \label{fig:mfeatFouriermcr}
    \end{subfigure}
    \begin{subfigure}{0.49\linewidth}
      \centering
        \includegraphics[width=\linewidth]{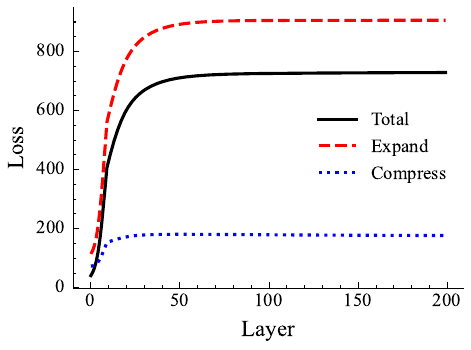}
        \caption{ESS-ReduNet}
        \label{fig:mfeatFouriermcr2}
    \end{subfigure}
  \caption{mfeatFourier}
\end{figure}

\begin{figure}[htb]
    \begin{subfigure}{0.49\linewidth}
      \centering
        \includegraphics[width=\linewidth]{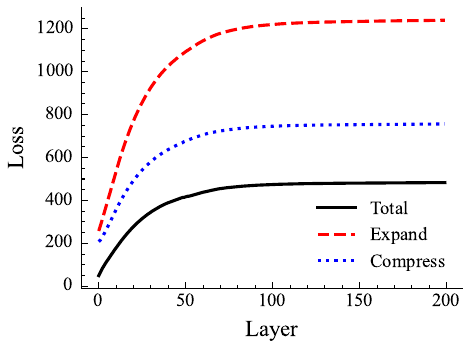}
        \caption{ReduNet}
        \label{fig:muskmcr}
    \end{subfigure}
    \begin{subfigure}{0.49\linewidth}
      \centering
        \includegraphics[width=\linewidth]{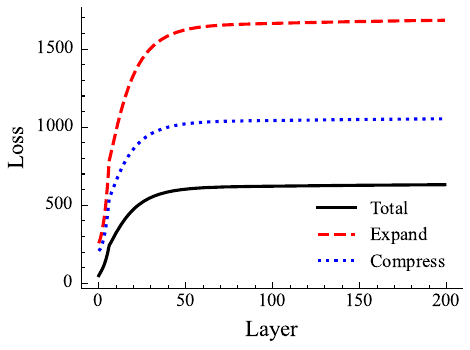}
        \caption{ESS-ReduNet}
        \label{fig:muskmcr2}
    \end{subfigure}
  \caption{musk}
  \label{fig:mcrmusk}
\end{figure}

Figures \ref{fig:mcrcovtype} to \ref{fig:mcrmusk} 
depict the curve of objective functions in six data sets, where ESS-ReduNet achieves higher values of MCR$^2$.

\subsection{Rank Trend}
\begin{figure}[htb]
    \begin{subfigure}{0.49\linewidth}
      \centering
        \includegraphics[width=\linewidth]{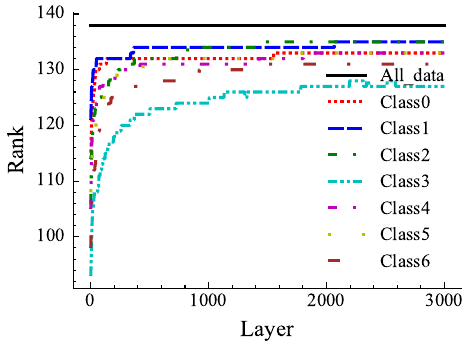}
        \caption{ReduNet}
        \label{fig:covtyperank}
    \end{subfigure}
    \begin{subfigure}{0.49\linewidth}
      \centering
        \includegraphics[width=\linewidth]{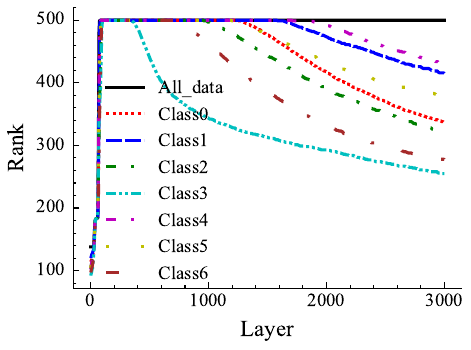}
        \caption{ESS-ReduNet}
        \label{fig:covtyperank2}
    \end{subfigure}
  \caption{covtype}
  \label{fig:rankcovtype}
\end{figure}

\begin{figure}[htb]
    \begin{subfigure}{0.49\linewidth}
      \centering
        \includegraphics[width=\linewidth]{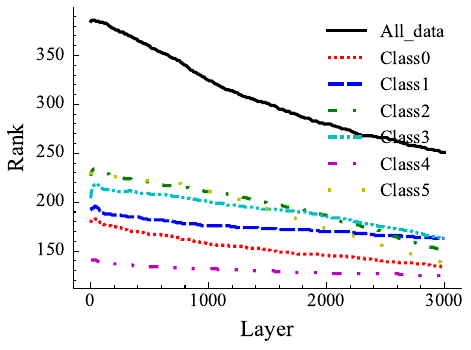}
        \caption{ReduNet}
        \label{fig:gasrank}
    \end{subfigure}
    \begin{subfigure}{0.49\linewidth}
      \centering
        \includegraphics[width=\linewidth]{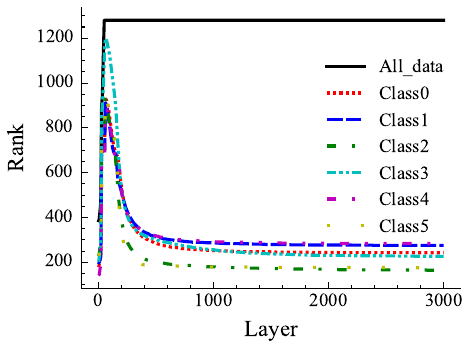}
        \caption{ESS-ReduNet}
        \label{fig:gasrank2}
    \end{subfigure}
  \caption{gas}
\end{figure}

\begin{figure}[htb]
    \begin{subfigure}{0.49\linewidth}
      \centering
        \includegraphics[width=\linewidth]{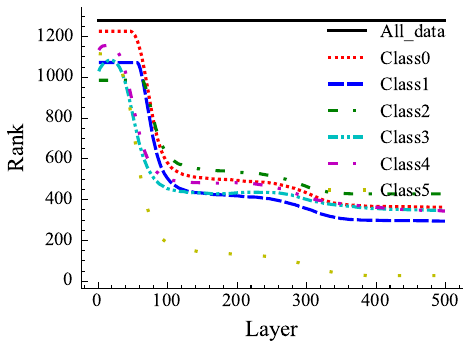}
        \caption{ReduNet}
        \label{fig:HARv1rank}
    \end{subfigure}
    \begin{subfigure}{0.49\linewidth}
      \centering
        \includegraphics[width=\linewidth]{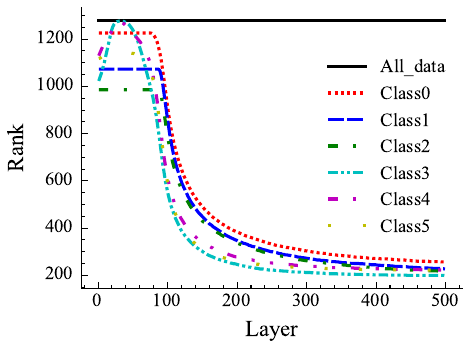}
        \caption{ESS-ReduNet}
        \label{fig:HARv1rank2}
    \end{subfigure}
  \caption{HARv1}
\end{figure}

\begin{figure}[htb]
    \begin{subfigure}{0.49\linewidth}
      \centering
        \includegraphics[width=\linewidth]{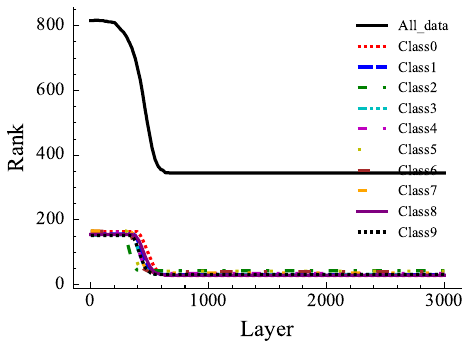}
        \caption{ReduNet}
        \label{fig:mfeatFactorsrank}
    \end{subfigure}
    \begin{subfigure}{0.49\linewidth}
      \centering
        \includegraphics[width=\linewidth]{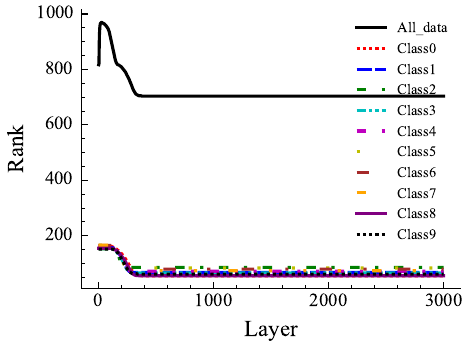}
        \caption{ESS-ReduNet}
        \label{fig:mfeatFactorsrank2}
    \end{subfigure}
  \caption{mfeatFactors}
\end{figure}

\begin{figure}[htb]
    \begin{subfigure}{0.49\linewidth}
      \centering
        \includegraphics[width=\linewidth]{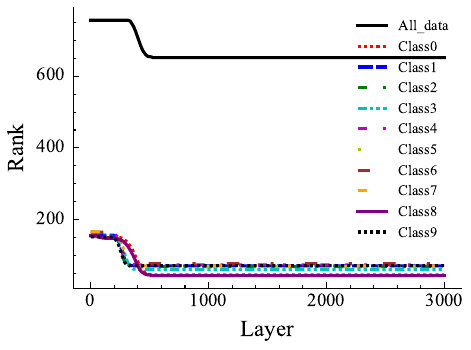}
        \caption{ReduNet}
        \label{fig:mfeatFourierrank}
    \end{subfigure}
    \begin{subfigure}{0.49\linewidth}
      \centering
        \includegraphics[width=\linewidth]{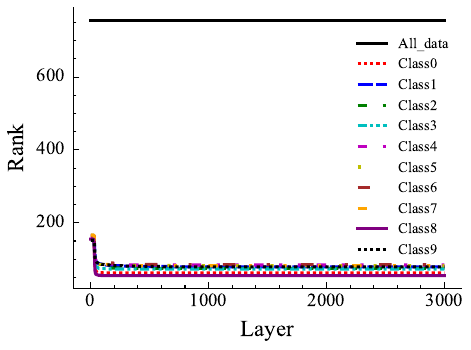}
        \caption{ESS-ReduNet}
        \label{fig:mfeatFourierrank2}
    \end{subfigure}
  \caption{mfeatFourier}
\end{figure}

\begin{figure}[htb]
    \begin{subfigure}{0.49\linewidth}
      \centering
        \includegraphics[width=\linewidth]{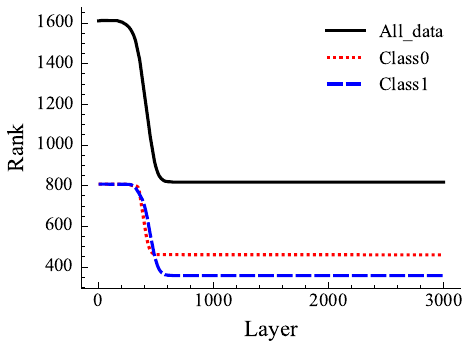}
        \caption{ReduNet}
        \label{fig:muskrank}
    \end{subfigure}
    \begin{subfigure}{0.49\linewidth}
      \centering
        \includegraphics[width=\linewidth]{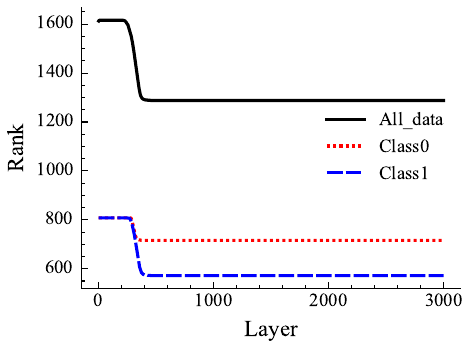}
        \caption{ESS-ReduNet}
        \label{fig:muskrank2}
    \end{subfigure}
  \caption{musk}
  \label{fig:rankmusk}
\end{figure}

Figures \ref{fig:rankcovtype} to \ref{fig:rankmusk} show the changes in rank across six datasets. It is evident that the features transformed by ESS-ReduNet have a larger overall spanned space. 
Moreover, there is no occurrence of the overall spanned space collapsing, as depicted in Figure \ref{fig:gasrank}.

\subsection{Condition Number}
\begin{figure}[htb]
    \begin{subfigure}{0.49\linewidth}
      \centering
        \includegraphics[width=\linewidth]{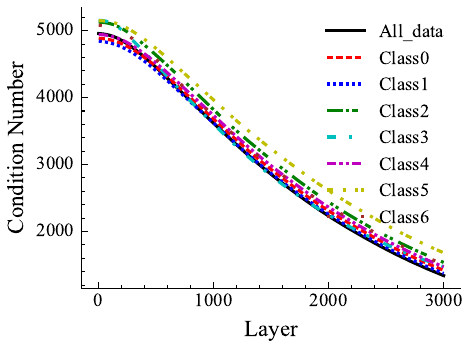}
        \caption{ReduNet}
        \label{fig:covtypecn}
    \end{subfigure}
    \begin{subfigure}{0.49\linewidth}
      \centering
        \includegraphics[width=\linewidth]{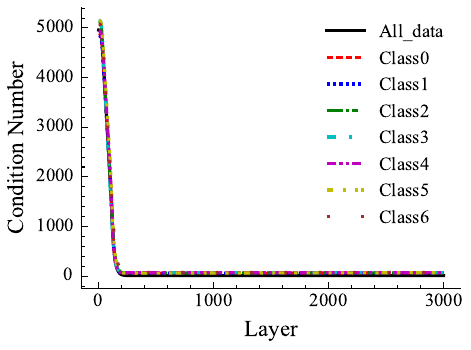}
        \caption{ESS-ReduNet}
        \label{fig:covtypecn2}
    \end{subfigure}
  \caption{covtype}
  \label{fig:cncovtype}
\end{figure}

\begin{figure}[htb]
    \begin{subfigure}{0.49\linewidth}
      \centering
        \includegraphics[width=\linewidth]{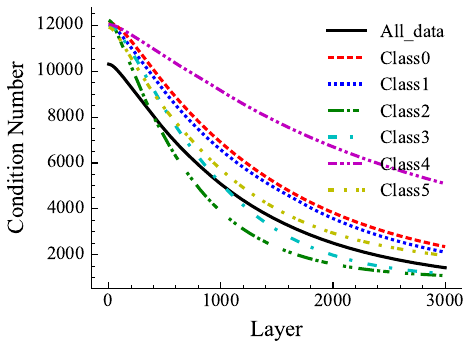}
        \caption{ReduNet}
        \label{fig:gascn}
    \end{subfigure}
    \begin{subfigure}{0.49\linewidth}
      \centering
        \includegraphics[width=\linewidth]{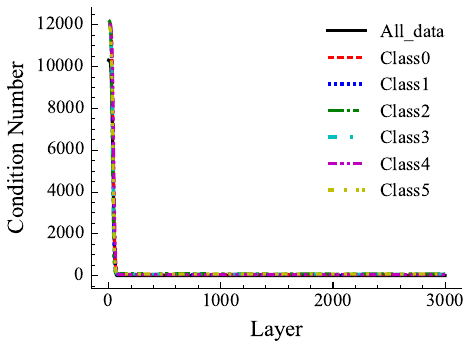}
        \caption{ESS-ReduNet}
        \label{fig:gascn2}
    \end{subfigure}
  \caption{gas}
\end{figure}

\begin{figure}[htb]
    \begin{subfigure}{0.49\linewidth}
      \centering
        \includegraphics[width=\linewidth]{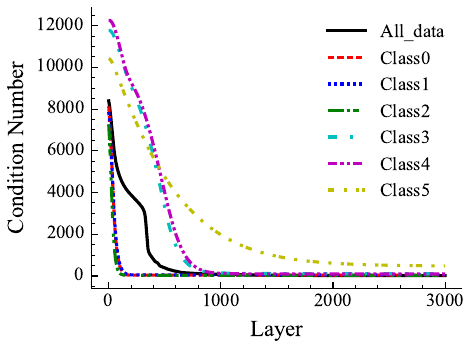}
        \caption{ReduNet}
        \label{fig:HARv1cn}
    \end{subfigure}
    \begin{subfigure}{0.49\linewidth}
      \centering
        \includegraphics[width=\linewidth]{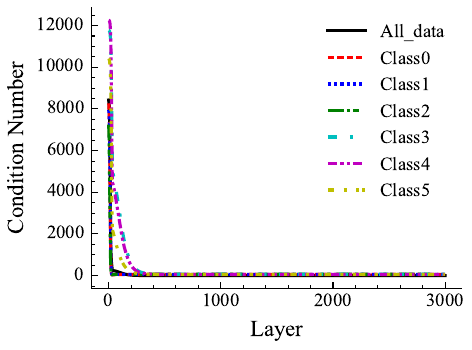}
        \caption{ESS-ReduNet}
        \label{fig:HARv1cn2}
    \end{subfigure}
  \caption{HARv1}
\end{figure}

\begin{figure}[htb]
    \begin{subfigure}{0.49\linewidth}
      \centering
        \includegraphics[width=\linewidth]{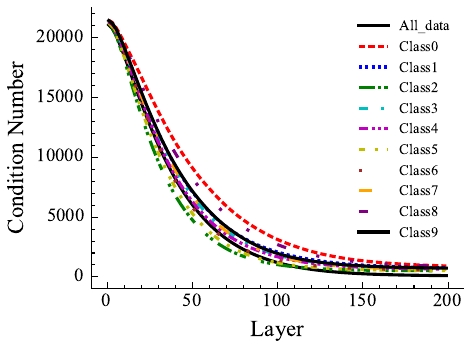}
        \caption{ReduNet}
        \label{fig:mfeatFactorscn}
    \end{subfigure}
    \begin{subfigure}{0.49\linewidth}
      \centering
        \includegraphics[width=\linewidth]{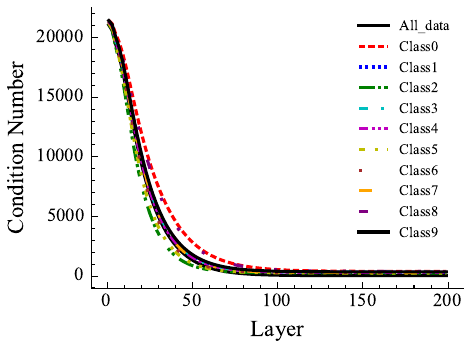}
        \caption{ESS-ReduNet}
        \label{fig:mfeatFactorscn2}
    \end{subfigure}
  \caption{mfeatFactors}
\end{figure}

\begin{figure}[htb]
    \begin{subfigure}{0.49\linewidth}
      \centering
        \includegraphics[width=\linewidth]{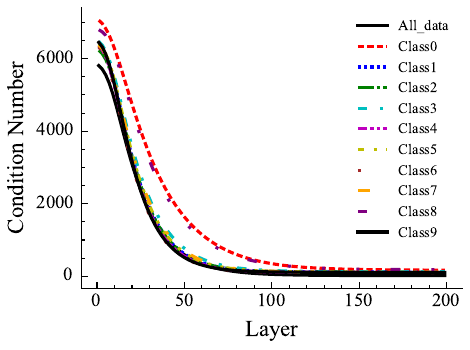}
        \caption{ReduNet}
        \label{fig:mfeatFouriercn}
    \end{subfigure}
    \begin{subfigure}{0.49\linewidth}
      \centering
        \includegraphics[width=\linewidth]{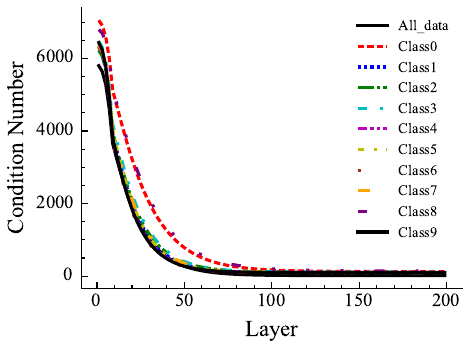}
        \caption{ESS-ReduNet}
        \label{fig:mfeatFouriercn2}
    \end{subfigure}
  \caption{mfeatFourier}
\end{figure}

\begin{figure}[htb]
    \begin{subfigure}{0.49\linewidth}
      \centering
        \includegraphics[width=\linewidth]{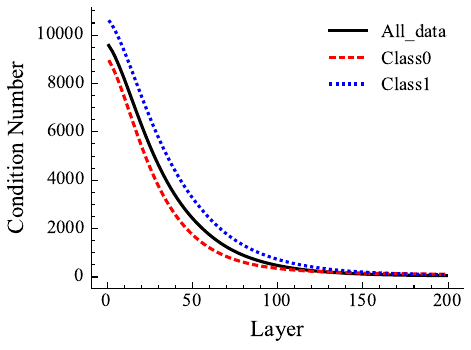}
        \caption{ReduNet}
        \label{fig:muskcn}
    \end{subfigure}
    \begin{subfigure}{0.49\linewidth}
      \centering
        \includegraphics[width=\linewidth]{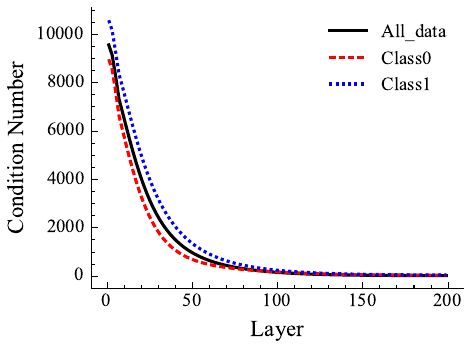}
        \caption{ESS-ReduNet}
        \label{fig:muskcn2}
    \end{subfigure}
  \caption{musk}
  \label{fig:cnmusk}
\end{figure}

Figures \ref{fig:cncovtype} to \ref{fig:cnmusk} display the changes in the condition numbers across six datasets. It is apparent that ESS-ReduNet achieves lower condition numbers more rapidly, indicating that the entire linear system stabilizes more quickly.
\end{document}